%% file: main.tex
\documentclass[10pt]{article}
\usepackage[T1]{fontenc}
\usepackage{lmodern}
\usepackage{arxiv}
\usepackage{natbib}
\setcitestyle{authoryear,round,citesep={;},aysep={,},yysep={;}}
\renewcommand{\shorttitle}{\textit{Rotation in Equally Spaced Overcomplete ICA}}
\renewcommand{\headeright}{Preprint}
\date{}
\usepackage{amsmath,amssymb,amsthm,graphicx,booktabs,array}
\usepackage[hidelinks,bookmarksnumbered]{hyperref}
\usepackage{url}

\newtheorem{theorem}{Theorem}
\newtheorem{proposition}{Proposition}
\newtheorem{corollary}{Corollary}
\newtheorem{lemma}{Lemma}

\newcommand{\E}{\mathbb E}
\newcommand{\R}{\mathbb R}
\newcommand{\C}{\mathbb C}
\newcommand{\Z}{\mathbb Z}
\newcommand{\cE}{\mathcal E}
\newcommand{\cH}{\mathcal H}
\newcommand{\cU}{\mathcal U}
\newcommand{\TV}{\operatorname{TV}}
\newcommand{\He}{\mathrm{He}}
\newcommand{\cum}{\operatorname{cum}}
\newcommand{\up}{u^{\perp}}
\newcommand{\norm}[1]{\lVert#1\rVert}
\newcommand{\Fnorm}[1]{\lVert#1\rVert_{\mathrm F}}

\title{Identifiable but Hard to Estimate:\\ Rotation in Equally Spaced Overcomplete ICA\thanks{ChatGPT assisted with manuscript organization, language editing, proof development, and checks of proofs and numerical calculations.}}
\input{authors}
\hypersetup{
  pdftitle={Identifiable but Hard to Estimate: Rotation in Equally Spaced Overcomplete ICA},
  pdfauthor={\authornames}
}

\begin{document}
\maketitle
\input{abstract}
\input{introduction}
\input{body}
\begingroup
\setlength{\bibsep}{2pt}

\input{references}
\endgroup

\appendix
\input{ext_s1}
\input{phase_details}

\input{ext_s2}
\input{ext_s3}
\input{ext_s4}
\end{document}

%% file: authors.tex
\newcommand{\authornames}{Wenjun Xiao; Yinghao Fang; Hanxin Chen; Sihan Jia; Vince D. Calhoun; Yuda Bi}
\author{%
  Wenjun Xiao$^{1}$\thanks{Wenjun Xiao and Yinghao Fang contributed equally to this work and share first authorship.} \quad Yinghao Fang$^{2}$\footnotemark[2] \quad Hanxin Chen$^{3}$\\[3pt]
  \bfseries Sihan Jia$^{4}$ \quad Vince D. Calhoun$^{5}$ \quad Yuda Bi$^{4}$\thanks{Corresponding author: \href{mailto:yudabitrends@gmail.com}{\texttt{yudabitrends@gmail.com}}.}\\[6pt]
  \normalfont\small $^{1}$The George Washington University\\
  \normalfont\small $^{2}$University of Health and Rehabilitation Sciences\\
  \normalfont\small $^{3}$University of California, San Diego \quad $^{4}$Georgia State University\\
  \normalfont\small $^{5}$Georgia Institute of Technology
}

%% file: abstract.tex
\begin{abstract}
How many observations are needed to estimate a common rotation in two-dimensional overcomplete independent component analysis? We study equally spaced mixing columns with known source count, common source law, and column norm. For a fixed non-Gaussian fourth-order Gram--Charlier source law, the rotation is identifiable modulo the column spacing. Yet the sample size for fixed relative angular accuracy grows exponentially with the source count. We characterize this sample size, up to absolute constants, by the reciprocal of the squared Hellinger distance between two rotations separated by half the column spacing. An explicit radially weighted phase estimator attains this scale uniformly over the rotation family, using the observation radius to prevent cancellation in the mean of an angular harmonic. We also determine the exact exponential decay rate of single-observation total variation for the same pair. This gives bounds on the sample size exponent; its exact value remains open. Finally, a noisy binary-source example shows how an exact zero line of the observed characteristic function can have an exponentially small normal derivative.
\end{abstract}

%% file: introduction.tex
\section{Introduction}
Consider ten independent sources mixed into two observed coordinates along equally spaced directions. This is an overcomplete independent component analysis (ICA) model: there are more sources than observed coordinates. The source count, common source law, and column norm are known, leaving the common rotation angle as the only unknown parameter. A Gram--Charlier source law of excess kurtosis four makes the angle identifiable modulo the $18^\circ$ spacing. Suppose the angular error must be strictly below $4.5^\circ$, with probability at least $75\%$ at every rotation. Numerical evaluation of the theoretical lower bound \eqref{eq:hellinger-sample} gives approximately $60.3$ million observations for this guarantee. Every unit projection has excess kurtosis $0.6$, so the difficulty persists despite visible non-Gaussianity.

This example separates two questions about ICA. Population identifiability asks whether the observation law determines the mixing parameter, up to the model's unavoidable ambiguities. Statistical estimation asks how many observations are needed to distinguish parameters at a specified accuracy and confidence. Non-Gaussianity is central to classical ICA \citep{comon1994,hyvarinenoja2000}, but its presence alone does not quantify the second question. In the opening example, fourth moments reveal non-Gaussianity while carrying no information about the common rotation.

We study this distinction in a family that isolates the effect of the mixing geometry. The observed dimension stays equal to two, the source law is fixed as the source count $q$ grows, and the columns have norm $\sqrt{2/q}$. The observation covariance is therefore the identity at every rotation. All model ingredients except the common angle are supplied to the estimator. A lower bound in this setting describes difficulty that remains even with that information; it does not rely on having to learn the source law or separate several unknown mixing parameters.

The accuracy scale is also part of the question. Rotating by one column spacing leaves the observation law unchanged, so the target is an angle modulo $\pi/q$. We require error strictly below one quarter of that spacing, with success probability at least $3/4$ at every rotation. This keeps the required resolution fixed relative to the identifiable parameter range as the columns become more closely spaced. The results below concern this uniform, fixed-relative-accuracy target.

\subsection{Results and mechanism}
For a known Gram--Charlier source law, Theorem~\ref{thm:hellinger} characterizes the required sample size up to absolute constants for every source count. A test between two rotations separated by half the column spacing gives the lower bound: it is proportional to the reciprocal of their single-observation squared Hellinger distance. A radially weighted phase estimator achieves the same scale uniformly over the full rotation family. Thus, for this model and loss, the difficulty of estimating a continuous parameter is captured by the separation of one explicit pair of observation laws.

The connection between the lower and upper bounds comes from the angular structure of the density. Equal spacing cancels all nonconstant angular frequencies below $2q$. For Gram--Charlier sources, the remaining nonconstant terms have frequencies $2q$ and $4q$, and the phase of the first identifies the rotation. Its amplitude depends on radius and can change sign, so unweighted averaging can cancel contributions that carry the same rotation information. Weights computed from the known model align these contributions. The radial distribution itself is independent of the rotation, but the observed radius indicates how to use the angular component. The second moment of the conditional harmonic amplitude then controls both the estimator's error and the Hellinger separation.

Theorem~\ref{thm:barrier} quantifies how the observation laws approach one another as the source count grows. For the same half-spacing pair, it gives the exact exponential decay rate of single-observation total variation (TV). Combined with Theorem~\ref{thm:hellinger}, this yields sample-size bounds that are exponential in $q$, despite the fixed observed dimension and the one-dimensional unknown parameter. The exact sample-size exponent remains open: total variation depends on the first absolute radial moment of the harmonic amplitude, whereas squared Hellinger distance is comparable to its second moment. The exact TV rate therefore provides exponent bounds, rather than an exact asymptotic formula for the required sample size.

A complementary binary-source example examines an exact zero line of the observed characteristic function. Keeping the equally spaced geometry, we show that the normal derivative along such a line can be exponentially small. This describes the sensitivity of a population feature, and motivates the local analyses in the appendices. The global rotation sample-complexity theorem and these local moment-based analyses have different targets and assumptions; Table~\ref{tab:statistical-targets} records the distinctions.

\subsection{Related work}\label{sec:related}
\paragraph{ICA identifiability and recovery.}
Conditions for population identifiability, separability, and uniqueness in linear ICA are studied by \citet{erikssonkoivunen}; \citet{wang} treats identifiability in overcomplete models. These results concern what can be determined from the observation law. Recovery guarantees additionally depend on the source and mixing assumptions. Fourier PCA provides finite-sample recovery guarantees under quantitative conditions on tensorized mixing columns and source moments \citep{gvx}. The semidefinite approach studies recovery through the subspace spanned by rank-one column outer products, with a population analysis for random directions \citep{podosinnikova}. For square mixing, polynomial-time recovery with unknown Gaussian noise is possible under source and conditioning assumptions \citep{arora2012,belkin2013}. Our family has a known source law and known relative column directions, leaving only a common rotation to estimate. Its lower bound concerns this restricted target, while its matching upper bound uses the supplied model information.

\paragraph{An earlier information-theoretic lower bound.}
\citet[Theorem~4]{anderson} construct ICA pairs in fixed dimension $d$ from disjoint subsets of $4k^2$ random directions. The sources follow instance-dependent scaled Poisson laws, and the observations include Gaussian noise with polynomially bounded variance. The laws are exponentially close in $(k/\log k)^{1/d}$. Their theorem gives a testing lower bound. Here the observed dimension is two, the common source law is fixed as $q$ grows, and there is no additive observation noise. We determine an exact TV exponent for the equally spaced family and construct an estimator matching its Hellinger scale. These are different geometric constructions and asymptotic parameters, rather than directly comparable bounds for the same model.

\paragraph{Moment information and statistical separation.}
In high-noise multireference alignment, informative moment orders govern sample complexity \citep{perry2019,bandeiramra2019}; estimation under group actions relates such rates to invariant polynomials \citep{bandeiraorbit}. The connection here is the loss of low-order information through symmetry. Our varying parameter is the source count at fixed observed covariance, rather than the noise level in an alignment model. Related distinctions between identifiability and estimation rates arise in finite mixtures: strong and weak identifiability control how distributional error translates into parameter error \citep{honguyenstrong2016,honguyenweak2016}, and local minimax rates depend on overfitting \citep{heinrichkahn2018}. In the present ICA family, the source law remains fixed and the changing mixing geometry makes separated rotations difficult to distinguish.

\paragraph{Organization.}
Sections~\ref{sec:rotation}--\ref{sec:barrier} establish the rotation bounds and explain the roles of radial weighting and symmetry. Section~\ref{sec:binary} studies sensitivity at a characteristic-function zero. Section~\ref{sec:discussion} discusses the unresolved sample-size exponent, departures from equal spacing, and learning the source law. Appendix~\ref{app:s1} contains the rotation proofs, while Appendices~\ref{app:s2}--\ref{app:s4} develop the complementary local estimation analyses.

%% file: body.tex
\section{Rotation identifiability and sample complexity}\label{sec:rotation}
Let $q\ge2$ be known. Write $u(\theta)=(\cos\theta,\sin\theta)^T$ and $\up(\theta)=(-\sin\theta,\cos\theta)^T$. We observe $n$ independent draws of $X=A_\theta S$, where
\[
 A_\theta=\sqrt{2/q}\,[u(\theta),u(\theta+\pi/q),\ldots,u(\theta+(q-1)\pi/q)]
\]
and $S=(s_0,\ldots,s_{q-1})^T$ has independent coordinates with the same symmetric, unit-variance law $f$. The source law, spacing, and common column norm are known. The only unknown parameter is the common rotation $\theta$. The normalization gives $A_\theta A_\theta^T=I_2$. Write $P_\theta$ for the law of $X$ and $p_\theta$ for its density when it exists.

A rotation by $\pi/q$ permutes the columns up to sign. Because the sources are symmetric and identically distributed, this leaves the observation law unchanged. We therefore estimate the rotation modulo one column spacing, using the distance
\[
 d_q(\theta,\theta')=\min_{k\in\Z}|\theta-\theta'-k\pi/q|,\qquad \varrho=\pi/(2q).
\]
Thus $\varrho$ is the largest separation on the parameter circle.

For a given source law, let $n^*(q)$ be the smallest sample size for which an estimator succeeds with probability at least $3/4$ for every $\theta$. Success means $d_q(\hat\theta,\theta)<\pi/(4q)$. The target is one quarter of the column spacing. Equivalently, success requires circular error strictly below $\pi/2$ in the phase coordinate $2q\theta\pmod{2\pi}$. This fixed phase accuracy makes the statistical target comparable across source counts, even as the columns become more closely spaced. Set $n^*(q)=\infty$ if the guarantee is impossible. The model is overcomplete for $q>2$ and includes the square case $q=2$.

\subsection{Which rotations have different observation laws?}
Covariance alone cannot distinguish the rotations: it is $I_2$ at every angle. With standard Gaussian sources, even the full observation law is the same, $\mathcal N(0,I_2)$. To obtain an identifiable family, consider the Gram--Charlier densities
\[
 \phi_a(s)=\phi(s)\big[1+a(s^4-6s^2+3)/24\big],\qquad 0<a\le4,
\]
where $\phi$ is the standard normal density and $a$ is the excess kurtosis. The polynomial $s^4-6s^2+3$ has minimum $-6$ and mean zero under the standard Gaussian law, so $0<a\le4$ gives nonnegative densities that integrate to one.

The density relative to the standard bivariate Gaussian is a polynomial of degree at most $4q$. Rotation by $\pi/q$ restricts its angular frequencies to multiples of $2q$, and reflection symmetry leaves cosine terms. Thus only $2q$ and $4q$ can be nonconstant angular frequencies. Rotating the mixing columns shifts their phases; we track the first harmonic to identify the rotation. In polar coordinates $x=r u(\psi)$,
\begin{equation}
 p_\theta(r u(\psi))=\phi_{\mathrm N,2}(r u(0))
 \big[A_0(r)+2A_1(r)\cos t_\theta+2A_2(r)\cos(2t_\theta)\big],
 \qquad t_\theta=2q(\psi-\theta).
 \label{eq:angular-density}
\end{equation}
Here $\phi_{\mathrm N,2}$ is the standard bivariate Gaussian density and the $A_j$ are real polynomials. In the bracket, $A_0(r)$ is the angular average, while $2A_1(r)$ and $2A_2(r)$ are the coefficients of the two cosine terms.

Lemma~\ref{s:lem:gc-density} shows that $A_1$ is nonzero. At any radius with $A_1(r)\ne0$, the coefficient of $e^{2iq\psi}$ in the bracket is $A_1(r)e^{-2iq\theta}$. Equality of observation laws implies equality of the continuous densities and hence of these coefficients. Therefore $P_\theta=P_{\theta'}$ if and only if $\theta=\theta'$ modulo $\pi/q$. The phase of this coefficient thus identifies the common rotation and provides the basis for the estimator below.

\subsection{The sample size for uniform rotation estimation}
We compare two rotations separated by half the column spacing. Their single-observation squared Hellinger distance is
\[
 H_q^2=\int_{\R^2}\big(\sqrt{p_0(x)}-\sqrt{p_\varrho(x)}\big)^2\,dx.
\]
This convention has no factor of $1/2$. Two-point testing is a standard route to estimation lower bounds \citep{yu1997,tsybakov}; Hellinger moduli also connect testing and estimation in other models \citep{donoholiu1991}. Here the half-spacing distance governs estimation over the full parameter circle.

\begin{theorem}[Uniform rotation estimation]\label{thm:hellinger}
Let $q\ge2$ and let the common source density be $\phi_a$ for a known $a\in(0,4]$. Then
\begin{equation}
 \frac{1}{4H_q^2}\ \le\ n^*(q)\ \le\
 \left\lceil\frac{64}{H_q^2}\right\rceil .
 \label{eq:hellinger-sample}
\end{equation}
\end{theorem}

Theorem~\ref{thm:hellinger} relates estimation over the full parameter circle to testing two half-spacing rotations. It determines $n^*(q)$ up to universal multiplicative constants, whose ratio is $256$ apart from integer rounding.

The estimator weights the observed harmonic by its conditional amplitude. Define
\[
 B(r)=\frac{2A_1(r)}{A_0(r)},\qquad M_q=\E[B(R)^2],\qquad
 \hat\theta=\frac{1}{2q}\arg\!\left[\frac1n\sum_{i=1}^n B(R_i)e^{2iq\psi_i}\right]
 \pmod{\pi/q}.
\]
Here $R=\|X\|$, and $(R_i,\psi_i)$ are the polar coordinates of observation $X_i$. Values at $R_i=0$ or a zero average are arbitrary. Weighting by $B(R)$ aligns the conditional harmonic means across radii and prevents sign cancellation in their average. The amplitude second moment $M_q$ controls both the estimator's concentration and the Hellinger separation.

\begin{proof}
\emph{Radial weights.}
The radial law is common to all rotations. Lemma~\ref{s:lem:gc-density} gives $A_2(r)>0$ for $r>0$. At $t_\theta=\pi/2$, nonnegativity in \eqref{eq:angular-density} then gives $A_0(r)\ge2A_2(r)>0$. Let $R=\|X\|$. For $r>0$, set $B(r)=2A_1(r)/A_0(r)$ and $C(r)=2A_2(r)/A_0(r)$. The conditional density of $\psi\in[0,2\pi)$ given $R=r$ is
\[
 (1+B(r)\cos t_\theta+C(r)\cos(2t_\theta))/(2\pi).
\]
We have $0\le C\le1$, and evaluating this density at $t_\theta=0,\pi$ yields $|B|\le1+C\le2$.

The radial law has positive density on $(0,\infty)$ and $A_1$ is a nonzero polynomial. Thus $M_q=\E[B(R)^2]>0$.

\emph{Half-spacing comparison.}
The half-spacing rotation changes the sign of the $2q$ harmonic and leaves the $4q$ harmonic unchanged. Fix $r$ and write $t=2q\psi$. Separate the unchanged part, $U=1+C(r)\cos(2t)$, from the sign-changing part, $V=B(r)\cos t$. The two conditional densities are then $(U+V)/(2\pi)$ and $(U-V)/(2\pi)$. Rationalizing the difference of their square roots gives
\[
 \frac{V^2}{U}
 \le \big(\sqrt{U+V}-\sqrt{U-V}\big)^2
 \le \frac{2V^2}{U},
\]
where the quotients are taken to be zero when $U=0$, since then $V=0$. The bounds $\cos^2t\le U\le2$ follow from $0\le C\le1$. Integrating with respect to $d\psi/(2\pi)$ and the common radial law yields
\begin{equation}
 M_q/4\le H_q^2\le 2M_q.
 \label{eq:hellinger-weight}
\end{equation}

\emph{Lower bound.}
The strict success regions around $0$ and $\varrho$ are disjoint. Deciding for $0$ exactly when the estimate lies in the success region around $0$ yields a test with each error probability at most $1/4$. Thus $\TV(P_0^{\otimes n},P_\varrho^{\otimes n})\ge1/2$. Here $\TV(P,Q)=\sup_E|P(E)-Q(E)|$ is total variation.

The Hellinger convention above gives $\TV(P,Q)\le H(P,Q)$. For the product laws,
\[
 H^2(P_0^{\otimes n},P_\varrho^{\otimes n})=2[1-(1-H_q^2/2)^n]\le nH_q^2.
\]
Combining these inequalities gives the lower bound in \eqref{eq:hellinger-sample}.

\emph{Upper bound.}
Use $Z=B(R)e^{2iq\psi}$ and take $\hat\theta=\arg(\bar Z)/(2q)$ modulo $\pi/q$, where $\bar Z$ is the sample mean. Assign an arbitrary value if $\bar Z=0$. The conditional density gives $\E_\theta Z=(M_q/2)e^{2iq\theta}$ and $\E_\theta|Z|^2=M_q$.

The sample mean satisfies $\E_\theta|\bar Z-\E_\theta Z|^2\le M_q/n$. Chebyshev's inequality therefore bounds the probability of
$|\bar Z-\E_\theta Z|>|\E_\theta Z|/\sqrt2$ by $8/(nM_q)$. For $n\ge32/M_q$, this probability is at most $1/4$, uniformly over $\theta$. On the complementary event the phase error is at most $\pi/4$, giving $d_q(\hat\theta,\theta)\le\pi/(8q)$. By \eqref{eq:hellinger-weight}, $\lceil64/H_q^2\rceil$ observations suffice.
\end{proof}

\begin{figure}[!htbp]
 \centering
 \includegraphics[width=\linewidth]{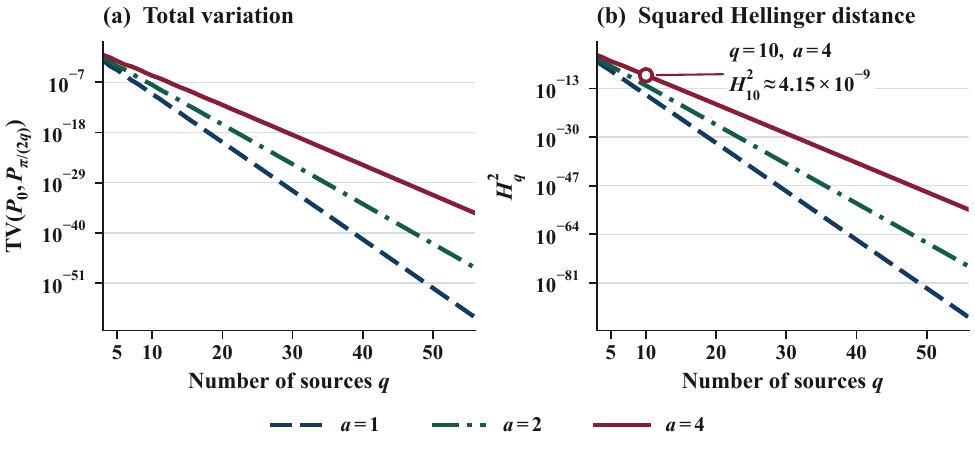}
 \caption{Single-observation distances between half-spacing rotations. Left: total variation (Theorem~\ref{thm:barrier}). Right: squared Hellinger distance (Theorem~\ref{thm:hellinger}). The highlighted point has $a=4$ and $q=10$. Numerical evaluation of the theoretical lower bound gives approximately $60.3$ million observations. The requirement is angular error strictly below $4.5^\circ$ with probability at least $3/4$ at every rotation. Both vertical axes are logarithmic.}
 \label{fig:rotation}
\end{figure}

At $a=4$ and $q=10$, radial and Cartesian calculations give $H_{10}^2\simeq4.14519\times10^{-9}$. Numerical evaluation of the theoretical lower bound in \eqref{eq:hellinger-sample} gives approximately $60.3$ million observations. Appendix~\ref{app:numerical-distances} compares the two integration methods and reports numerical refinement checks. Figure~\ref{fig:rotation} shows finite-$q$ distances; Section~\ref{sec:sample-exponent} studies their exponential rates.

\paragraph{Computing the estimator.}
Given $q$ and $a$, construct $A_0$ and $A_1$ from the Hermite representation in Lemma~\ref{s:lem:gc-density}. Evaluate $B(R_i)$ for each observation, form the weighted phase average displayed above, and reduce its argument divided by $2q$ modulo $\pi/q$.

\subsection{What the radial weights do}
At radius $r$, the conditional mean of $e^{2iq\psi}$ is $(B(r)/2)e^{2iq\theta}$. Without weights, the coefficient of $e^{2iq\theta}$ in the mean is $\E[B(R)]/2$, which can be small if the contributions at different radii cancel. A negative conditional amplitude reverses the mean direction of the harmonic even though the mixing rotation is the same.

After weighting by $B(R)$, the mean is $(\E[B(R)^2]/2)e^{2iq\theta}$ and the second absolute moment is $\E[B(R)^2]$. The weights correct the sign and emphasize radii with stronger harmonics. Radius alone carries no rotation information, because its law is common to all rotations.

\begin{figure}[!htbp]
 \centering
 \includegraphics[width=\linewidth]{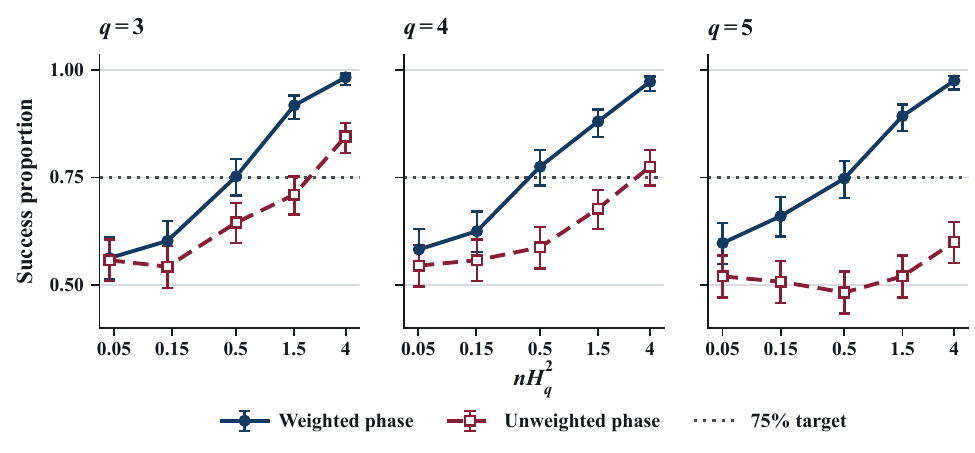}
 \caption{Success of weighted and unweighted phase estimation at $a=4$.
 The horizontal axis shows $nH_q^2$ on a logarithmic scale. A uniform random guess succeeds with probability $1/2$, because the success region covers half the parameter circle. The estimation target is success probability $3/4$.
 The error bars are pointwise $95\%$ Wilson intervals. Section~\ref{app:phase} gives the simulation protocol.}
 \label{fig:phase}
\end{figure}

Figure~\ref{fig:phase} illustrates the effect of weighting at $a=4$ and $q=3,4,5$, where the population unweighted mean is a positive multiple of $e^{2iq\theta}$. Rotation equivariance lets us fix $\theta=0$. The two estimators use the same observations within each replicate, and the success event is the one defining $n^*(q)$. Near $nH_q^2=0.5$, the weighted estimator succeeds in $75$--$78\%$ of trials across the three settings. The corresponding range for the unweighted estimator is $48$--$65\%$.

\section{Why rotation is hard to estimate}\label{sec:barrier}
\subsection{What symmetry cancels}
Cumulant tensors of order below $2q$ are rotation invariant whenever they exist. Odd orders vanish by symmetry. At even order $k$, the sum $\sum_j(a_j^Tt)^k$ contains only angular frequencies that are multiples of $2q$ as a function of $\theta$, where $a_j$ are the columns of $A_\theta$. Below the first permitted frequency, only the constant angular term can remain.

For $q\ge3$ and unit $t$, the fourth cumulant of $t^TX$ is $3a/(2q)$ under $\phi_a$, giving $0.6$ in the opening example. The zero-lag fourth cumulant used in tensor-based ICA \citep{cardoso91,foobi} therefore contains no rotation information here. Methods using temporal diversity, such as FOBIUM \citep{fobiumtsp}, exploit additional structure absent from this independent-sample model.

Under $\phi_a$, cumulants above order two are nonzero only at multiples of four. The first cumulant tensor that depends on the rotation is therefore of order $2q$ for even $q$ and $2q+2$ for odd $q$, as shown in Appendix~\ref{app:s1}. The observation density has lowest nonconstant angular frequency $2q$ in both cases.

\subsection{The exact total variation exponent}
In high-noise multireference alignment, sample complexity depends on informative moment orders \citep{perry2019,bandeiramra2019}, which can change with the shift distribution \citep{abbe2019}. The group-action framework connects these rates to separating invariant degrees \citep{bandeiraorbit}. In our model, increasing the source count raises the informative angular frequency to $2q$ and reduces each column norm to $\sqrt{2/q}$. The next theorem quantifies the resulting separation of the full observation laws for the pair used in Theorem~\ref{thm:hellinger}.

\begin{theorem}[Exact total variation exponent]\label{thm:barrier}
Fix $a\in(0,4]$ in the common rotation model with Gram--Charlier sources. With $\varrho=\pi/(2q)$ and the constant $\varkappa_a>0$ defined by the one-dimensional maximization in \eqref{eq:tv-exponent-definition},
\[
 \lim_{q\to\infty}-\frac{1}{q}\log\TV(P_0,P_\varrho)=\varkappa_a .
\]
\end{theorem}

For $a=4$, $\varkappa_4\simeq1.5288$. The general expression is $\varkappa_a=-\max_{v>0}e_a(v)$. The function $e_a(v)$ describes the exponential rate of the surviving angular harmonic's amplitude at frequency radius $v\sqrt{q/2}$, after its oscillatory factor has been separated. Equation~\eqref{eq:tv-exponent-definition} gives this rate explicitly by factoring the characteristic function on a circle in frequency space. Maximizing over the radius identifies the slowest single-observation TV decay.

A real-frequency evaluation gives a lower bound on total variation. The oscillatory factor can vanish at individual radii. The proof avoids these zeros by choosing a nearby radius where the amplitude has the same exponential rate. For the upper bound, Plancherel's identity controls the density difference on a ball through an integrated squared difference in frequency space. A tail estimate handles the region outside the ball. These two bounds have the same exponential rate. Appendix~\ref{app:s1}, Proposition~\ref{s:prop:exact}, gives the bounds for finite source counts and the limiting argument.

\subsection{What this gives for the sample size exponent}\label{sec:sample-exponent}
Integrating the absolute difference between the two conditional angle densities gives $\TV(P_0,P_\varrho)=(2/\pi)\E|B(R)|$. The squared Hellinger distance is comparable to $\E[B(R)^2]$ by \eqref{eq:hellinger-weight}. Thus the exact TV rate describes a first absolute moment, while the sample size depends on the rate of the second moment.

Using $\TV^2\le H_q^2\le 2\TV$ in Theorems~\ref{thm:hellinger} and~\ref{thm:barrier} gives a lower exponent of $\varkappa_a$ and an upper exponent of $2\varkappa_a$.

The finite-$q$ TV bounds in Appendix~\ref{app:s1} give $\log n^*(q)\ge\varkappa_a q-O(\log q)$. For $a=4$, numerical evaluation of a proved complex-frequency bound improves the upper exponent from $2\varkappa_4\simeq3.0577$ to approximately $2.3689$. Appendix~\ref{app:complex-bound} gives the auxiliary estimator, the exact bound at fixed parameters, and the parameter optimization. The exact limiting sample exponent remains open; the conjecture is stated there.

\subsection{A total variation bound for other symmetric source laws}\label{sec:general-tv}
The symmetry argument extends beyond the Gram--Charlier family, even when the observation density does not have a finite angular expansion. To control all surviving harmonics, we impose a growth condition on the source transform. Let the common symmetric unit-variance source law $f$ satisfy $\E e^{r|s|}<\infty$ for every $r>0$. Define its growth relative to the Gaussian transform by
\[
 \mu_f=\sup_{z\in\C\setminus\{0\}}
 \frac{\log\big|e^{-z^2/2}\E e^{zs}\big|}{|z|^2},\qquad s\sim f .
\]
If $\mu_f<1/2$ and $q\ge1/(1-2\mu_f)$, Proposition~\ref{s:prop:lower} gives
\begin{equation}
 \TV(P_0,P_\varrho)\le4\sqrt{\frac{2q}{1-2\mu_f}}\,(2\mu_f)^q .
 \label{eq:general-tv}
\end{equation}
At the half-spacing rotation, only the odd multiples of angular frequency $2q$ contribute to the density difference. A complex rotation of the Gaussian-normalized transform bounds the weighted sum of squared Hermite coefficients at these frequencies. This controls the density difference beyond the finitely many cumulants that agree, complementing finite-order moment matching arguments for Gaussian mixtures \citep{wuyang}.

The condition on $\mu_f$ controls source tails and characteristic-function decay as well as moment existence. It holds for $\phi_a$ and the Gaussian-smoothed binary law below (Appendix~\ref{app:s1}). For the larger source class, \eqref{eq:general-tv} establishes closeness of the observation laws; identifiability and a matching estimation upper bound require separate arguments.

\section{Sensitivity at a characteristic-function zero}\label{sec:binary}
Characteristic-function ICA uses derivatives \citep{yeredor2000,comonrajih2006}, including in two-sensor overcomplete identification \citep{taleb2001}. We examine the lines of zeros produced by source characteristic functions. Each line is normal to the corresponding mixing column. The question is how rapidly the observed characteristic function changes across an exact zero line.

The Gram--Charlier characteristic function has no real zeros. We therefore keep the equally spaced geometry and use binary sources with optional Gaussian smoothing: $s_j=bB_j+\tau Z_j$. The $B_j$ are equiprobable signs and the $Z_j$ are standard normals, all mutually independent. We take $b>0$ and $\tau\ge0$, with $b^2+\tau^2=1$ to give unit source variance. When $\tau>0$, the transform growth parameter satisfies $2\mu_f=b^2<1$, so \eqref{eq:general-tv} applies for sufficiently large $q$. The observed characteristic function (CF) is
\begin{equation}
 \Phi_x(\omega)=\E e^{i\omega^TX}
 = e^{-\tau^2\|\omega\|^2/2}\prod_{j=0}^{q-1}\cos(b\,a_j^T\omega).
 \label{eq:binary-cf}
\end{equation}
The first positive zero line associated with $a_0$ is $b\,a_0^T\omega=\pi/2$. It lies at distance $\rho_0=\pi\sqrt q/(2\sqrt2\,b)$ from the origin. Because the characteristic function is identically zero along this line, its tangential derivative is zero. Its gradient therefore measures sensitivity in the normal direction.

On the line, differentiation can only contribute through the vanishing cosine factor. The other cosines have magnitude at most one, and the Gaussian factor is largest at the point nearest the origin. Consequently
\begin{equation}
 \|\nabla_\omega\Phi_x(\omega)\|
 \le b\sqrt{2/q}\,\exp\!\left(-\frac{\pi^2\tau^2q}{16b^2}\right).
 \label{eq:binaryslope}
\end{equation}
Bound \eqref{eq:binaryslope} captures the Gaussian attenuation. The cosine product causes further decay and provides most of the decay exponent at 10 dB, where $b^2/\tau^2=10$. To separate these contributions, record the normal slope as a function of position along the line. Let
\[
 S_q(\nu)=\|\nabla_\omega\Phi_x(\rho_0u(\theta)+\nu\up(\theta))\|,\qquad
 \gamma_{\mathrm{bin}}=-\frac{1}{\pi}\int_0^\pi
      \log\cos\!\left(\frac{\pi}{2}\cos\varphi\right)d\varphi .
\]
The calculation in Appendix~\ref{s:binary-slope-rate} shows that at the point nearest the origin,
\[
 \lim_{q\to\infty}q^{-1}\log S_q(0)
     =-\gamma_{\mathrm{bin}}-\frac{\pi^2\tau^2}{16b^2}.
\]
The same limit holds for the maximum normal slope on any fixed bounded segment centered at that point. The binary product contributes $\gamma_{\mathrm{bin}}\simeq1.5061$ to the decay exponent per source, and this contribution remains when $\tau=0$. At $b^2/\tau^2=10$ (10 dB), the Gaussian contribution is about $0.0617$ per source.

An angular search uses a different derivative because changing the candidate angle moves the evaluation point in frequency space. Hold the observation law fixed and vary the candidate angle in $\rho_0u(\theta')+\nu\up(\theta')$, keeping $\rho_0$ and $\nu$ fixed. At $\theta'=\theta$, only the normal component of the frequency displacement contributes, so
\[
 \left|\left.\frac{d}{d\theta'}
 \Phi_x(\rho_0u(\theta')+\nu\up(\theta'))
 \right|_{\theta'=\theta}\right|=|\nu|S_q(\nu).
\]
It vanishes at $\nu=0$. Figure~\ref{fig:estimation}(a) in Appendix~\ref{app:s2} compares the angular derivative at $\nu=1$ with \eqref{eq:binaryslope}; at this offset, its magnitude equals the normal slope $S_q(1)$. The proved exponential rate applies at $\nu=0$ and to the maximum over a fixed bounded segment; rates at individual nonzero offsets may differ.

Appendix~\ref{app:s2} studies local estimation using a known real zero of the target source CF and an isolating angle--offset window, fixed or supplied independently of the estimation sample. With finitely many frequencies, identification additionally requires the population criterion to vanish only at the target. At fixed model size and grid, local variance depends on moment derivatives and the full covariance of sampled CF values; slopes alone do not establish a minimax rate.

\section{Discussion}\label{sec:discussion}
The rotation model separates identifiability from the amount of information available in a sample. The phase of the $2q$ harmonic identifies the common rotation, while the radial second moment of its conditional amplitude determines the sample size. Radius alone is uninformative about rotation, yet it tells the estimator how to combine angular observations without cancellation. Theorem~\ref{thm:hellinger} makes this distinction quantitative through a guarantee uniform over the parameter circle, at fixed accuracy relative to the column spacing. The guarantee uses the known source law and geometry; its matching upper bound does not establish the same performance when either must also be learned. The construction is explicit, but the theorem does not bound the numerical precision or computational cost needed to evaluate the radial weights as $q$ grows.

\paragraph{The unresolved sample-size exponent.}
The exact total variation exponent leaves a specific statistical question unresolved. Total variation measures the first absolute moment of the conditional harmonic amplitude, whereas squared Hellinger distance is comparable to its second moment. Determining the decay of the former therefore does not determine the sample-size exponent. The complex-frequency bound narrows the available upper bound, but the existence and value of the limiting sample-size exponent remain open. The conjecture in Appendix~\ref{app:complex-bound} asks whether optimizing that construction loses no exponential factor relative to the radial estimator. Resolving it would identify the exponent of a sample size already characterized, for each $q$, up to absolute constants. The finite-range fits support investigation of this question without establishing an asymptotic limit.

\paragraph{Departures from equal spacing.}
The role of symmetry also raises questions about how the difficulty changes when the geometry is perturbed.

Departures from equal spacing can make the covariance informative. Let the known offsets be $\alpha_j=j\pi/q+\epsilon_j$, and take mixing columns $\sqrt{2/q}\,u(\theta+\alpha_j)$. Their second angular component is $z_2=(2/q)\sum_j e^{2i\alpha_j}$, which gives
$\E[(X_1+iX_2)^2]=e^{2i\theta}z_2$. When $z_2\ne0$, the covariance identifies $\theta$ modulo $\pi$. For the Gram--Charlier family, estimate this expectation by the sample mean of $(X_1+iX_2)^2$. Divide the sample mean by the known $z_2$ and halve its phase to estimate $\theta$ modulo $\pi$. This estimator achieves angular accuracy $\varepsilon$ at fixed confidence using $O(|z_2|^{-2}\varepsilon^{-2})$ observations. The statistic's variance is bounded by $\E\|X\|^4$, uniformly bounded under this normalization. Indeed, the covariance $\Sigma$ has trace two and $\E\|X\|^4=(\operatorname{tr}\Sigma)^2+2\operatorname{tr}(\Sigma^2)+4a/q$.

This calculation identifies a mechanism by which known perturbations can restore information at a low moment order. Its usefulness depends on the size of $z_2$: a nonzero value establishes identification through covariance, but the stated covariance-based sample bound deteriorates as that value approaches zero. The calculation does not characterize the transition between covariance-based estimation and the higher angular harmonics near equal spacing.

For a larger model with unknown offsets, the two-point lower bound transfers when the loss restricts to $d_q$ on the exactly spaced subclass, under the same accuracy and success requirements. If $\theta$ remains a separate target parameter, the parameterization must distinguish the common rotation from a common shift of all offsets.

The transferred bound is a worst-case statement for a model containing that subclass. It does not say that every perturbed configuration is equally difficult, or supply an estimator for jointly unknown directions. Studying such models requires specifying which geometric features are known and how the target is represented before comparing rates.

\paragraph{Learning the source law.}
In the equally spaced model, $\E\|X\|^4=8+4a/q$, so the radial law identifies $a$ even when it is unknown. Whether an estimate of $a$ can be used in the radial weights while preserving the sample size guarantee remains open.

The issue is how error in the estimated source parameter affects the weighted harmonic when its mean is small. Identifying $a$ from radial observations does not by itself control this effect. More generally, the total variation bound for other symmetric source laws establishes closeness of observation laws under its transform-growth condition; identifiability and a matching upper bound still require separate arguments.

\paragraph{Local sensitivity and global estimation.}
The characteristic-function zero example addresses a complementary local question. An exact zero line identifies a geometric feature of the population transform, while its normal slope can be exponentially small. The local estimation calculations additionally require an isolating search window and, for a finite grid, a unique population minimum. The resulting estimator covariance depends on both moment derivatives and the joint fluctuations of the sampled transform values. A slope calculation alone therefore gives neither a global minimax rate nor a guarantee of locating the correct line. Choosing informative frequencies, accounting for their covariance, and obtaining bounds uniform in growing model size are natural next steps. These questions connect the local calculations to the uniform rotation problem while preserving their different assumptions and limits.

%% file: ext_s1.tex
\section{Proofs for the rotation model}\label{app:s1}

The proofs begin with the angular density used in Theorem~\ref{thm:hellinger}. The Hermite representation
converts transform differences into density differences, while the angular decomposition isolates the
terms that distinguish the two rotations. Together they give the general total-variation bound.
For Gram--Charlier sources, factoring the characteristic function on a circle gives the exact
TV exponent in Theorem~\ref{thm:barrier} and the sample-size exponent bounds in
Section~\ref{sec:sample-exponent}. Section~\ref{s:binary-slope-rate} treats the normal derivative in the
binary-source example of Section~\ref{sec:binary}.

\subsection{Notation and transform representation}
Let $q\ge2$. The independent sources have the same symmetric, unit-variance law $f$ with
$\E e^{r|s|}<\infty$ for all $r>0$. Write the mixing matrix as
\[
 A_\theta=r_q\,[u(\alpha_0),\ldots,u(\alpha_{q-1})],\qquad r_q=\sqrt{2/q},
\]
with $\alpha_j=\theta+j\pi/q$ and columns $a_j=r_q u(\alpha_j)$. Since $\sum_ju(\alpha_j)u(\alpha_j)^T=(q/2)I_2$ for $q\ge2$,
$A_\theta A_\theta^T=I_2$. A rotation by $\pi/q$ permutes the columns up to sign, so the distance between configurations is $d_q(\theta,\theta')=\operatorname{dist}(\theta-\theta',(\pi/q)\Z)$. We write
$\varrho=\pi/(2q)$ for the largest such distance. The comparison below uses $A_0$ and $A_\varrho$.

Let $P_\theta$ denote the law of $X=A_\theta S$ and $p_\theta$ its
density when it exists. Write $\phi_{\mathrm N,d}$ for the standard normal density on $\R^d$ and $\cum_k(s)$ for the
$k$th cumulant of $s$. The characteristic function of $X$ is denoted by $\Phi_x$; lowercase $x$ also denotes a realized value or a function argument.

For a symmetric source let $m_f(u)=e^{-u^2/2}\E e^{us}$, an even entire function with
$m_f(u)=1+O(u^4)$. For $\phi_a$, $m_f(u)=1+au^4/24$. Since $A_\theta A_\theta^T=I_2$, independence gives
\begin{equation}
 \E e^{t^TX}=e^{t^Tt/2}P_A(t),\qquad P_A(t)=\prod\nolimits_{j=0}^{q-1}m_f(a_j^Tt).
 \label{s:PB}
\end{equation}
The factor $P_A$ carries the dependence on the source law and mixing matrix. Under the
finiteness condition below, the Hermite transform of $P_A$ gives the density ratio relative to the standard
Gaussian density.
For a multi-index $\alpha=(\alpha_1,\alpha_2)$, write $t^\alpha=t_1^{\alpha_1}t_2^{\alpha_2}$ and $\alpha!=\alpha_1!\alpha_2!$.
Let $\cH$ map $t^\alpha$ to $\He_\alpha(x)=\He_{\alpha_1}(x_1)\He_{\alpha_2}(x_2)$, where $\He_k$ denotes the probabilists' Hermite polynomial.
For $P=\sum_\alpha p_\alpha t^\alpha$, define the squared Fischer norm by $\Fnorm P^2=\sum_\alpha\alpha!|p_\alpha|^2$.
Hermite expansions also describe local non-Gaussian geometry in ICA \citep{cardoso2003}.
The next lemma identifies the squared Fischer norm with the Gaussian $L^2$ norm of the Hermite
series and with an average over complex Gaussian arguments.

\begin{lemma}[Hermite representation]\label{s:lem:hermite}
If $\Fnorm{P_A}<\infty$, then $X$ has density $\phi_{\mathrm N,2}r_A$, where $r_A=\cH P_A\in L^2(\phi_{\mathrm N,2})$.
For two matrices satisfying this condition,
$\int(r_A-r_{A'})^2\phi_{\mathrm N,2}=\Fnorm{P_A-P_{A'}}^2$. Let $W$ have independent circular complex Gaussian entries with $\E|W_i|^2=1$. Then
$\Fnorm P^2=\E|P(W)|^2$ for every entire $P$, with both sides allowed to be infinite.
\end{lemma}

\begin{proof}
Let $p_\alpha$ be the Taylor coefficients of $P_A$. The generating function of the Hermite polynomials,
$e^{t^Tx-t^Tt/2}=\sum_\alpha\He_\alpha(x)t^\alpha/\alpha!$, and dominated convergence give
$\E\He_\alpha(X)=\alpha!\,p_\alpha$. Since $\He_k(y)=\E(y+iZ)^k$ for $Z\sim\mathcal N(0,1)$, the series is
dominated by
\[
 \sum\nolimits_\alpha|\He_\alpha(x)||t^\alpha|/\alpha!\le\prod\nolimits_i\E e^{|t_i|(|x_i|+|Z_i|)},
\]
which is integrable because $\E e^{c\norm X}<\infty$ for all $c>0$.

We next construct the candidate density from these Hermite moments. Since $\sum_\alpha\alpha!|p_\alpha|^2<\infty$, the
series $\tilde r=\sum_\alpha p_\alpha\He_\alpha$ converges in $L^2(\phi_{\mathrm N,2})$ and $\int\He_\alpha\tilde
r\phi_{\mathrm N,2}=\alpha!\,p_\alpha$. Cauchy--Schwarz gives
$\int e^{c\norm x}|\tilde r(x)|\phi_{\mathrm N,2}(x)\,dx\le\|\tilde r\|_{L^2(\phi_{\mathrm N,2})}(\E e^{2c\norm G})^{1/2}<\infty$ for $G\sim\mathcal N(0,I_2)$.
Thus the finite signed measure $\nu=P_\theta-\tilde r\phi_{\mathrm N,2}$ has all exponential moments and vanishing polynomial moments.
Its Laplace transform is entire and all its Taylor coefficients vanish, so $\nu=0$. Hence the candidate density represents the observation law.

Orthogonality of the $\He_\alpha$ gives the $L^2$ identity. For the complex Gaussian representation,
$\E[W^\alpha\bar W^{\alpha'}]=\alpha!\,\delta_{\alpha\alpha'}$.
Integrating over polydiscs and then applying monotone convergence gives
$\E|P(W)|^2=\sum_\alpha\alpha!|p_\alpha|^2$, the Fischer norm representation of \citet{bargmann}.
\end{proof}

\subsection{The Gram--Charlier angular density}
For the Gram--Charlier source, the transform factor is a finite polynomial. Its degree and symmetries
restrict the angular terms in the density. Its highest-degree part supplies a nonzero coefficient
that makes the density depend on the rotation.
\begin{lemma}[Angular density]\label{s:lem:gc-density}
For $q\ge2$ and $0<a\le4$, the observation density has the form \eqref{eq:angular-density}. Let
$K=(a/(96q^2))^q$. Then $A_2(r)=Kr^{4q}$ and the coefficient of $r^{4q}$ in $A_1(r)$ is $-4(-1)^q K$.
\end{lemma}

\begin{proof}
Here $P_{A_\theta}(t)=\prod_{j=0}^{q-1}[1+a(a_j^Tt)^4/24]$, a polynomial of degree $4q$.
The Hermite transform in Lemma~\ref{s:lem:hermite} is $\exp(-\Delta_t/2)$ on polynomials,
where $\Delta_t$ is the Laplacian. It commutes with rotations and reflections and preserves the highest-degree
homogeneous part. At $\theta=0$, rotation by $\pi/q$ and reflection about the first coordinate axis permute the unsigned
columns.
The rotation symmetry restricts angular frequencies to multiples of $2q$.
The degree bound excludes frequencies above $4q$, and reflection symmetry excludes sine terms.
Thus only the frequencies $0$, $2q$, and $4q$ remain. Rotating by $\theta$ gives the phase in \eqref{eq:angular-density}.

To compute the highest-degree coefficients, put $z=t_1+it_2$ for real $t$. Then
$a_j^Tt=(2q)^{-1/2}e^{-ij\pi/q}(z+e^{2ij\pi/q}\bar z)$ at $\theta=0$.
The product of the fourth powers of the phases equals one, and the product over roots of unity gives the top
homogeneous term $K[z^q-(-1)^q\bar z^q]^4$. The coefficients of $e^{4iq\psi}$ and $e^{2iq\psi}$
in this term are respectively $Kr^{4q}$ and $-4(-1)^qKr^{4q}$. Lower degrees cannot contribute to the
$4q$ harmonic or change the coefficient at degree $4q$ of the $2q$ harmonic.
\end{proof}

\subsection{Angular harmonics and factorization on a circle}
Decomposing the transform into angular frequencies makes the effect of rotation explicit.
We will bound the components that remain after subtracting the two configurations.
Write $\zeta=t_1+it_2$ and $\zeta'=t_1-it_2$, and expand an entire function as $P=\sum_{k,l}\tilde p_{kl}\zeta^k\zeta'^l$.
Define its angular component at frequency $\ell$ by $\Pi_\ell P=\sum_{k-l=\ell}\tilde p_{kl}\zeta^k\zeta'^l$.
For complex $t$, $\zeta$ and $\zeta'$ are separate coordinates and need not be complex conjugates.
For real $t=\norm tu(\psi)$, $\zeta^k\zeta'^l=\norm t^{k+l}e^{i(k-l)\psi}$.

To bound the high angular frequencies, we rescale the two complex coordinates in opposite directions.
For $\vartheta\in\R$ let $D_\vartheta$ act on
$\C^2$ by $(\zeta,\zeta')\mapsto(e^\vartheta\zeta,e^{-\vartheta}\zeta')$, a rotation by the imaginary angle
$-i\vartheta$. Part (c) below uses this rescaling to bound the sum of squared Fischer norms above a chosen angular frequency.

\begin{lemma}[Angular harmonics]\label{s:lem:harm}
(a) The monomials $\zeta^k\zeta'^l$ are orthogonal for the Fischer inner product, with
$\Fnorm{\zeta^k\zeta'^l}^2=2^{k+l}k!\,l!$. Hence $\Fnorm P^2=\sum_\ell\Fnorm{\Pi_\ell P}^2$.
\par(b) If $m_f$ is even, $\Pi_\ell P_{A_\theta}=0$ unless $2q$ divides $\ell$, and $\Pi_\ell
P_{A_{\theta+\psi}}=e^{-i\ell\psi}\Pi_\ell P_{A_\theta}$. In particular
$P_{A_0}-P_{A_\varrho}=2\sum_{m\ \mathrm{odd}}\Pi_{2qm}P_{A_0}$.
\par(c) For $\vartheta\ge0$ and every integer $\ell_0\ge0$,
$\sum_{\ell\ge\ell_0}\Fnorm{\Pi_\ell P}^2\le e^{-2\ell_0\vartheta}\E|P(D_\vartheta W)|^2$, and symmetrically for
$\ell\le-\ell_0$ with $D_{-\vartheta}$.
\end{lemma}

\begin{proof}
(a) The variables $\zeta=W_1+iW_2$ and $\zeta'=W_1-iW_2$ are independent circular complex Gaussians of variance $2$.
Thus $\E[\zeta^k\zeta'^l\overline{\zeta^{k'}\zeta'^{l'}}]=2^{k+l}k!\,l!\,\delta_{kk'}\delta_{ll'}$.
\par(b) Under $R_\psi$, the coordinates transform as $(\zeta,\zeta')\mapsto(e^{i\psi}\zeta,e^{-i\psi}\zeta')$.
Consequently, $\Pi_\ell(P\circ R_{-\psi})=e^{-i\ell\psi}\Pi_\ell P$.
Since $P_{A_{\theta+\psi}}=P_{A_\theta}\circ R_{-\psi}$, the stated phase relation follows.
Moreover $R_{\pi/q}a_j=a_{j+1}$ for $j<q-1$ and $R_{\pi/q}a_{q-1}=-a_0$, so $P_{A_\theta}$ is invariant under
$R_{\pi/q}$ when $m_f$ is even, and $\Pi_\ell P_{A_\theta}=0$ unless $e^{-i\ell\pi/q}=1$. For $\psi=\varrho$ and
$\ell=2qm$, $e^{-i\ell\varrho}=(-1)^m$.
\par(c) By (a), $\E|P(D_\vartheta W)|^2=\sum_\ell
e^{2\ell\vartheta}\Fnorm{\Pi_\ell P}^2\ge e^{2\ell_0\vartheta}\sum_{\ell\ge\ell_0}\Fnorm{\Pi_\ell P}^2$.
\end{proof}

For Gram--Charlier sources, factoring the source polynomial gives the angular dependence explicitly.
The resulting formula supplies both the amplitude for the exact TV exponent and the Fourier
coefficient for rotation estimation.
The exponential moment condition $\E e^{c\norm X}<\infty$ for all $c$ implies that $\Phi_x$ extends
to an entire function on $\C^2$. Symmetry of $X$ then gives $\E\cos(z^TX)=\Phi_x(z)$ for $z\in\C^2$. For $\phi_a$ the source CF is
$\chi(u)=e^{-u^2/2}(1+au^4/24)$, and on the circle $\omega=\tilde v\sqrt{q/2}\,u(\psi)$ we have
$a_j^T\omega=\tilde v\cos(\psi-\alpha_j)$.

\begin{lemma}[Circle factorization]\label{s:lem:circle}
Let the source law be $\phi_a$ and let $q\ge2$. Choose $\tilde v\in\C\setminus\{0\}$ with $|\arg\tilde v|<\pi/4$.
Using the principal fourth root, define $y_1=(a\tilde v^4/24)^{-1/4}e^{i\pi/4}$ and set $y_2=iy_1$.
For $k=1,2$, let $w_k$ be the unique root of $w^2-2y_kw+1=0$ inside the unit disc.
Set $U_k=w_k^{2q}$ and $\Psi=2q(\psi-\theta)$. Then
\begin{equation}
 \begin{split}
 &\Phi_x\big(\tilde v\sqrt{q/2}\,u(\psi)\big)\\
 &\quad=\frac{e^{-q\tilde v^2/4}\prod_{k=1,2}(1+U_k^2-2U_k\cos\Psi)}
 {[(1+w_1^2)(1+w_2^2)]^{2q}}.
 \end{split}
 \label{s:circleprod}
\end{equation}
For real $\tilde v=v>0$, $w_2=-\bar w_1$. With $w=w_1$, $U=w^{2q}$ and $\Lambda_0(v)=-v^2/4-4\log|1+w^2|$,
$\Phi_x=e^{q\Lambda_0(v)}|1+U^2-2U\cos\Psi|^2$, and
\begin{equation}
 \begin{split}
 &\Phi_x\big|_{\theta=0}-\Phi_x\big|_{\theta=\varrho}\\
 &\quad=-8e^{q\Lambda_0(v)}(1+|U|^2)\operatorname{Re}(U)\cos(2q\psi).
 \end{split}
 \label{s:circlediff}
\end{equation}
\end{lemma}

\begin{proof}
Put $c=a\tilde v^4/24$. The roots of $1+cy^4$ are $\pm y_1$ and $\pm y_2$, and $1+cy^4=\prod_{\pm,k}(1\mp y/y_k)$
because $y_1^2y_2^2=-y_1^4=1/c$. Since $|\arg\tilde v|<\pi/4$, $\arg y_1\in(0,\pi/2)$ and $\arg y_2\in(\pi/2,\pi)$,
so no root lies in $[-1,1]$. For such $y$, the two roots of $w^2-2yw+1=0$ are reciprocal; denote the one inside the unit disc by $w$.
The corresponding root for $-y$ is $-w$.

This choice of root rewrites each source factor in terms of angular exponentials.
Using $2yw=1+w^2$ gives $1-\cos x/y=(1-we^{ix})(1-we^{-ix})/(1+w^2)$. Hence
$(1-\cos x/y)(1+\cos x/y)=(1-w^2e^{2ix})(1-w^2e^{-2ix})/(1+w^2)^2$. For $x=\psi-\alpha_j$, we have $e^{2ix}=e^{2i(\psi-\theta)}e^{-2\pi ij/q}$.
Applying $\prod_{j=0}^{q-1}(1-ze^{-2\pi ij/q})=1-z^q$ to the two factors yields
$\prod_j(1-\cos^2(\psi-\alpha_j)/y^2)=(1+U^2-2U\cos\Psi)/(1+w^2)^{2q}$. The Gaussian factors give
$\prod_je^{-\tilde v^2\cos^2(\psi-\alpha_j)/2}=e^{-q\tilde v^2/4}$, since $\sum_je^{2i\alpha_j}=0$.

For real $v$,
$y_2=-\bar y_1$, so $w_2=-\bar w_1$ and the two factors are complex conjugate. Expanding
$|1+U^2-2U\cos\Psi|^2$, the shift $\theta\mapsto\theta+\varrho$ changes the sign of $\cos\Psi$ only, which gives
\eqref{s:circlediff}.
\end{proof}

On every circle $\Phi_x$ is thus a trigonometric polynomial with the frequencies $0$, $\pm2q$, $\pm4q$. Fourier
inversion preserves these angular frequencies, so $p_0-p_\varrho$ is a single harmonic of frequency $2q$.

The exact exponent proof also needs the root's phase to vary with the radius.
For real $v>0$,
$y_1=(24/(av^4))^{1/4}e^{i\pi/4}$, and the root $w=w(v)$ of $w^2-2y_1w+1=0$ inside the unit disc is an analytic
function of $y_1$ off $[-1,1]$. It is therefore real-analytic in $v>0$. As $v\to0$, $y_1\to\infty$ along the ray
$e^{i\pi/4}$ and $w\sim1/(2y_1)$, so $|w|\to0$ and $\arg w\to-\pi/4$. As $v\to\infty$, $y_1\to0$ and $w\to-i$. In
particular $\arg w$ is not constant.

\subsection{Cancellation of low-order cumulants}
The same rotational symmetry explains why low-order moments cannot distinguish the configurations.
The order-$k$ cumulant tensor of $X$ is $\cum_k(s)\sum_ja_j^{\otimes k}$, which vanishes for odd $k$ by source symmetry.
For $t=\norm tu(\psi)$, the geometric factor has the associated form
$T_k(t)=\sum_j(a_j^Tt)^k=r_q^k\norm t^k\sum_j\cos^k(\alpha_j-\psi)$.
For even $k$, expand $\cos^kx=2^{-k}\sum_{r=0}^k\binom kre^{i(k-2r)x}$.
The sum $\sum_{j=0}^{q-1}e^{i(k-2r)(\theta+j\pi/q-\psi)}$ vanishes unless $2q$ divides $k-2r$.
Hence $T_k$ does not depend on $\theta$ for even $k<2q$.
For $k=2q$ its $\theta$-dependent
part is $2^{1-2q}qr_q^{2q}\norm t^{2q}\cos(2q(\theta-\psi))$, so the order-$2q$ cumulants of $P_0$ and
$P_\varrho$ differ whenever $\cum_{2q}(s)\ne0$.

Equivalently, the two unit orientations of each of the $q$ lines form a spherical $(2q-1)$-design on the circle \citep{dgs}. The cancellation at order $k$ requires only that the symmetric source law have a finite $k$th
moment. Adding independent Gaussian noise with fixed covariance preserves the cancellation.

For the Gram--Charlier law, the nonzero cumulants above order two occur at multiples of four, by the series for
$\log(1+at^4/24)$. The first such multiple at least $2q$ is $2q$ for even $q$ and $2q+2$ for odd $q$.
The $\cos^k$ expansion at this order has a positive coefficient at frequency $2q$, which survives the sum.

\subsection{Exponential total variation bound}
The next bound combines the absence of low angular frequencies with a Gaussian growth envelope
for the transform over the entire complex plane.
Let $\mu_f=\sup_{u\in\C\setminus\{0\}}|u|^{-2}\log|m_f(u)|$. On the real and imaginary axes, this definition gives $\E e^{us}\le e^{(1/2+\mu_f)u^2}$ and
$|\E e^{iys}|\le e^{-(1/2-\mu_f)y^2}$, respectively. Thus $\mu_f<1/2$ requires both Gaussian-type tails and CF decay. The Gaussian law has $\mu_f=0$.

\begin{proposition}\label{s:prop:lower}
Let $q\ge2$ and suppose the common symmetric, unit-variance source law $f$ has all exponential
moments and satisfies $\mu_f<1/2$. For every $z\in[1,1/(2\mu_f))$ (with upper endpoint $+\infty$ if $\mu_f=0$),
\[
 \begin{aligned}
 \TV(P_0,P_\varrho)&\le\tfrac12\norm{r_{A_0}-r_{A_\varrho}}_{L^2(\phi_{\mathrm N,2})}\\
 &\le\frac{\sqrt2\,z^{-q}}{\sqrt{(1-2\mu_fz)(1-2\mu_f/z)}},
 \end{aligned}
\]
and if $q\ge1/(1-2\mu_f)$, then $\TV(P_0,P_\varrho)\le4\sqrt{2q/(1-2\mu_f)}\,(2\mu_f)^q$.
\end{proposition}

\begin{proof}
We first bound the transform after rescaling the two complex coordinates.
Choose $\vartheta\ge0$ such that $2\mu_f e^{2\vartheta}<1$; this condition is automatic when $\mu_f=0$. The variables $\zeta=W_1+iW_2$ and $\zeta'=W_1-iW_2$ are independent circular complex Gaussians with variance $2$.
The factors of $P_{A_0}(D_\vartheta W)$ are $m_f(u_j)$, where
$u_j=(r_q/2)(e^{\vartheta-i\alpha_j}\zeta+e^{-\vartheta+i\alpha_j}\zeta')$. By the definition of $\mu_f$,
$|P_{A_0}(D_\vartheta W)|^2\le\exp(2\mu_f\sum_j|u_j|^2)$, and because $\sum_je^{2i\alpha_j}=0$,
$\sum_j|u_j|^2=\frac12(e^{2\vartheta}|\zeta|^2+e^{-2\vartheta}|\zeta'|^2)$. Under the chosen restriction $2\mu_f e^{2\vartheta}<1$, the exponential moments of the independent variables $|\zeta|^2$ and $|\zeta'|^2$, each of mean $2$, give
$\E|P_{A_0}(D_{\pm\vartheta}W)|^2\le[(1-2\mu_fe^{2\vartheta})(1-2\mu_fe^{-2\vartheta})]^{-1}$.

Setting $\vartheta=0$ in this estimate verifies the integrability needed for the Hermite representation:
$\Fnorm{P_{A_0}}<\infty$. Lemma~\ref{s:lem:harm} gives
$\Fnorm{P_{A_\varrho}}=\Fnorm{P_{A_0}}$, so Lemma~\ref{s:lem:hermite} applies to both matrices. By Cauchy--Schwarz, Lemma~\ref{s:lem:hermite}, and Lemma~\ref{s:lem:harm}(b),
\[
 \TV\le\tfrac12\Fnorm{P_{A_0}-P_{A_\varrho}}
 =\Fnorm{\sum_{m\,\mathrm{odd}}\Pi_{2qm}P_{A_0}}.
\]
Apply Lemma~\ref{s:lem:harm}(c) with $\ell_0=2q$ to the positive and negative frequencies separately. The square of the last norm is at most
$e^{-4q\vartheta}(\E|P_{A_0}(D_\vartheta W)|^2+\E|P_{A_0}(D_{-\vartheta}W)|^2)$. Setting $z=e^{2\vartheta}$ gives the
first bound.

For the second bound, take $z=(1-1/q)/(2\mu_f)\ge1$ and use $1-2\mu_f/z\ge1-2\mu_f$ and
$(1-1/q)^{-q}\le4$. (If $\mu_f=0$, then $|m_f|\le1$ on $\C$, so $m_f\equiv1$, $f$ is Gaussian and $P_0=P_\varrho$.)
\end{proof}

\emph{Checking the growth condition.} For $\phi_a$, the inequality $|1+au^4/24|\le1+a|u|^4/24$ is attained when $u^4>0$.
With $x=\sqrt{a/24}|u|^2$, this gives $\mu_{\phi_a}=\mu_*\sqrt{a/24}$, where $\mu_*=\max_{x>0}x^{-1}\log(1+x^2)\simeq0.8047$.
In particular, $2\mu_{\phi_a}\le0.658$ for $a\le4$.

Gaussian smoothing and rescaling also connect the laws within the Gram--Charlier family.
For $a<4$, let $c_a^2=\sqrt a/2$ and take independent $s_4\sim\phi_4$ and $g\sim\mathcal N(0,1)$.
The variable $c_a s_4+\sqrt{1-c_a^2}\,g$ has law $\phi_a$, as follows from its moment generating
function $e^{u^2/2}(1+au^4/24)$. Because $A_\theta A_\theta^T=I_2$, the $\phi_a$ observation law is obtained by scaling the $\phi_4$ observations by $c_a$ and adding independent Gaussian noise with covariance $(1-c_a^2)I_2$.

For a second example satisfying the growth condition, take $s=\sqrt{1-\tau^2}\,s'+\tau g$, where $s'$ is an equiprobable sign, $g$ is an independent standard Gaussian, and $0<\tau<1$.
Then $m_f(u)=\cosh(\sqrt{1-\tau^2}u)e^{-(1-\tau^2)u^2/2}$.
For real $x,y$, the bounds $|\cosh(x+iy)|\le\cosh x$ and $\cosh x\le e^{x^2/2}$ give
\[
 |m_f(x+iy)|\le\cosh(\sqrt{1-\tau^2}x)e^{-(1-\tau^2)(x^2-y^2)/2}
 \le e^{(1-\tau^2)y^2/2}.
\]
Therefore $2\mu_f\le1-\tau^2<1$. Equality holds by taking $u=2\pi i/\sqrt{1-\tau^2}$. This source law also has real CF zeros at
$(k+1/2)\pi/\sqrt{1-\tau^2}$, $k\in\Z$, so the same model exhibits both the TV bound and the zero lines.

\subsection{The exact TV exponent for Gram--Charlier sources}
Use the scaled frequency radius $v$, root $w(v)$, and factor $\Lambda_0(v)$ from Lemma~\ref{s:lem:circle}.
The difference in \eqref{s:circlediff} separates an exponential amplitude from an oscillating cosine.
Let $e_a(v)$ be the amplitude's exponent at radius $v$, and let $\varkappa_a$ be the negative of its
maximum. Proposition~\ref{s:prop:exact} identifies $\varkappa_a$ as the exact TV decay exponent.
\begin{equation}
 \begin{aligned}
 e_a(v)&=\Lambda_0(v)+2\log|w(v)|,\\
 \varkappa_a&=-\max\nolimits_{v>0}e_a(v).
 \end{aligned}
 \label{eq:tv-exponent-definition}
\end{equation}
Since $1+w^2=2y_1w$, $e_a(v)=-v^2/4+\log(av^4/384)+2\log(1/|w(v)|)$. The maximum exists because $e_a$ is continuous on $(0,\infty)$ and tends to $-\infty$ at both endpoints.

\begin{proposition}\label{s:prop:exact}
For $\phi_a$, $a\in(0,4]$, and every $q\ge2$,
\[
 \begin{aligned}
 \TV(P_0,P_\varrho)&\ge4\max_{v>0}e^{qe_a(v)}|\cos(2q\arg w(v))|,\\
 \TV(P_0,P_\varrho)&\le\big(1+T_qe^{\varkappa_a}\sqrt{8qI_a}\big)e^{-\varkappa_aq},
 \end{aligned}
\]
with $I_a=\int_0^\infty ve^{2e_a(v)}dv$ and $T_q^2=(2+4\mu_{\phi_a})(\varkappa_aq+\log8)/\cos^2(\pi/8)$.
The constant $I_a$ is finite: $e_a(v)=2\log v+O(1)$ as $v\to0$ and $e_a(v)=-v^2/4+4\log v+O(1)$ as $v\to\infty$. Hence
$q^{-1}\log\TV(P_0,P_\varrho)\to-\varkappa_a$.
\end{proposition}

\begin{proof}
For real $\omega$, $|\Phi_0(\omega)-\Phi_\varrho(\omega)|\le2\TV$. At $\psi=0$, \eqref{s:circlediff} gives
$|\Phi_0-\Phi_\varrho|=8e^{q\Lambda_0(v)}(1+|U|^2)|\operatorname{Re}U|\ge8e^{qe_a(v)}|\cos(2q\arg w)|$.
Maximizing over $v$ gives the lower bound.

The cosine may vanish where $e_a$ is maximized. We therefore choose a nearby radius where its
absolute value is at least $1/2$. Let $\bar v_a$ maximize $e_a$ and fix $0<h<\bar v_a$. Since $\arg w$ is real-analytic and not constant, its range on $[\bar v_a-h,\bar v_a+h]$ is an interval of
some length $\ell_h>0$. Every interval on which $|\cos t|<1/2$ has length at most $\pi/3$.
Thus, if $2q\ell_h>\pi/3$, some $v\in[\bar v_a-h,\bar v_a+h]$ satisfies $|\cos(2q\arg w(v))|\ge1/2$.
The lower bound then gives $q^{-1}\log\TV\ge\min_{|v-\bar v_a|\le h}e_a(v)+q^{-1}\log2$.
Taking $q\to\infty$ and then $h\to0$ yields $\liminf q^{-1}\log\TV\ge-\varkappa_a$.
Combining this limit inferior with the TV upper bound in Proposition~\ref{s:prop:lower} gives
$\varkappa_a\ge-\log(2\mu_{\phi_a})>0$.

For the upper bound, we split the density integral into a tail and a bounded disc.
Let $\mu=\mu_{\phi_a}$ and $T>0$. The two configurations differ only by rotation, so $\norm X$ has the same law under both.
Thus $\frac12\int_{\norm x>T}|p_0-p_\varrho|\le\Pr_0\{\norm X>T\}$. The moment bound $\E e^{\lambda^TX}\le e^{(1/2+\mu)\norm\lambda^2}$ gives a Gaussian tail bound in every unit direction.
The eight directions $u(k\pi/4)$, $k=0,\ldots,7$, satisfy $\max_{0\le k\le7}u(k\pi/4)^Tx\ge\norm x\cos(\pi/8)$.
A union bound therefore gives $\Pr\{\norm X>T\}\le8\exp(-T^2\cos^2(\pi/8)/(2+4\mu))$.
At $T=T_q$, this upper bound equals $e^{-\varkappa_aq}$.

Inside the disc, we use the circle formula to bound the Fourier transform of the density difference.
By the Cauchy--Schwarz inequality and Plancherel's identity,
$\frac12\int_{\norm x\le T}|p_0-p_\varrho|\le T\norm{\Phi_0-\Phi_\varrho}_{L^2}/(4\sqrt\pi)$. In the coordinates $\omega=v\sqrt{q/2}\,u(\psi)$, the area element is $(q/2)v\,dv\,d\psi$.
Using \eqref{s:circlediff}, $\int_0^{2\pi}\cos^2(2q\psi)\,d\psi=\pi$, and $(1+|U|^2)^2(\operatorname{Re}U)^2\le4|w|^{4q}$ gives
\[
 \norm{\Phi_0-\Phi_\varrho}_{L^2}^2
 \le128\pi q\int_0^\infty ve^{2qe_a(v)}\,dv
 \le128\pi q e^{-2(q-1)\varkappa_a}I_a.
\]
Combining the disc and tail bounds gives the stated upper bound and $\limsup q^{-1}\log\TV\le-\varkappa_a$.
\end{proof}

For $a=4$, numerical evaluation gives $\varkappa_4\simeq1.5288$. If the observation is $Y=X+\varepsilon$ with independent $\varepsilon\sim\mathcal N(0,\sigma^2I_2)$, the same exponent formula applies with $a$ replaced by $a/(1+\sigma^2)^2$.
Indeed, $Y/\sqrt{1+\sigma^2}$ has the noiseless observation law with source density $\phi_{a/(1+\sigma^2)^2}$.

\subsection{Sample complexity of rotation estimation}\label{app:complex-bound}
The TV bound gives a necessary sample size. A phase estimator at complex frequencies gives a
sufficient sample size and hence an upper bound on the sample-size exponent.
\emph{Lower bound.} Since $d_q(0,\varrho)=\pi/(2q)$, the test that decides $\theta=0$ when
$d_q(\hat\theta,0)<\pi/(4q)$ errs under $\theta=0$ only if $d_q(\hat\theta,0)\ge\pi/(4q)$ and under $\theta=\varrho$
only if $d_q(\hat\theta,\varrho)\ge\pi/(4q)$. The sum of its two error probabilities is at least
$1-\TV(P_0^{\otimes n},P_\varrho^{\otimes n})\ge1-n\TV(P_0,P_\varrho)$ (Le Cam's two-point bound \citep{tsybakov}),
so success with probability $3/4$ at both rotations requires $n\ge1/(2\TV)$.
Proposition~\ref{s:prop:exact} then gives $\log n^*(q)\ge\varkappa_aq-O(\log q)$.

Theorem~\ref{thm:hellinger} already determines the sample size up to absolute constants at each fixed $q$ through $H_q^2$.
The construction below gives an explicit exponent bound as $q$ grows.

\emph{Upper bound with a known source law.} To obtain an explicit exponential bound in $q$,
we estimate the rotation from the phase of one angular Fourier coefficient at complex frequencies.
Write $\tilde v=ve^{i\xi}$ with radius parameter $v>0$ and phase $\xi$ satisfying $|\xi|<\pi/4$.
The parameter $\vartheta$ below controls a contour shift; $\theta$ remains the unknown rotation.
The angular average below extracts the Fourier coefficient whose phase records the rotation.
Let $N_{\rm ang}=4q$, $\varphi_k=\pi k/N_{\rm ang}$ and
$\tilde\omega_k=\tilde v\sqrt{q/2}\,u(\varphi_k)\in\C^2$. Define
\[
 T_{\rm cf}(x)=\frac1{N_{\rm ang}}\sum_{k=0}^{N_{\rm ang}-1}\cos(\tilde\omega_k^Tx)e^{2iq\varphi_k},
 \qquad \hat T=\frac1n\sum_{i=1}^n T_{\rm cf}(X_i).
\]
Using $w_k$ and $U_k$ from Lemma~\ref{s:lem:circle}, define the coefficient at zero rotation by
\[
 m_q=-[U_1(1+U_2^2)+U_2(1+U_1^2)]e^{-q\tilde v^2/4}[(1+w_1^2)(1+w_2^2)]^{-2q}.
\]
When $m_q\ne0$, set $\hat\theta=(2q)^{-1}\arg(\hat T/m_q)$ modulo $\pi/q$.
Assign an arbitrary value when $\hat T=0$. Let
$\tilde\Lambda(u)=\Lambda_0(u)+u^2/2$.
The proposition compares the signal mean with the statistic's second moment to control phase error
uniformly over rotations. Part (d) makes the exponent explicit when one root contribution dominates the other.

\begin{proposition}\label{s:prop:complex}
Let the source be $\phi_a$, $a\in(0,4]$.
\par(a) $\E_\theta\hat T=m_qe^{2iq\theta}$ for every $\theta$.
\par(b) For every $\vartheta>0$, uniformly over $\theta$,
\[
 \E_\theta|T_{\rm cf}(X)|^2\le\bar V_q
 :=16(1-e^{-4q\vartheta})^{-2}\exp\{q[-4\vartheta+\tilde\Lambda(2vL_{\xi,\vartheta})]\},
\]
where $L_{\xi,\vartheta}=(\sin^2\xi\cosh^2\vartheta+\cos^2\xi\sinh^2\vartheta)^{1/2}$.
\par(c) If $m_q\ne0$ and $n\ge8\bar V_q/|m_q|^2$, then
$\Pr_\theta\{d_q(\hat\theta,\theta)\le\pi/(8q)\}\ge3/4$ for every $\theta$.
\par(d) Suppose $|w_1|>|w_2|$, $|U_2/U_1|\le1/8$, and $|U_1|\le1/2$.
Then $\bar V_q/|m_q|^2\le64(1-e^{-4q\vartheta})^{-2}e^{q\Gamma_a(v,\xi,\vartheta)}$, where
\[
 \begin{aligned}
 \Gamma_a(v,\xi,\vartheta)&=-4\vartheta+\tilde\Lambda(2vL_{\xi,\vartheta})-2\ell_\xi,\\
 \ell_\xi&=-\operatorname{Re}(\tilde v^2)/4-2\log|1+w_1^2|-2\log|1+w_2^2|+2\log|w_1|.
 \end{aligned}
\]
\end{proposition}

\begin{proof}
\emph{Signal mean.}
(a) The mean $\E\cos(\tilde\omega_k^TX)=\Phi_x(\tilde\omega_k)$ is given by \eqref{s:circleprod} at $\psi=\varphi_k$.
As a function of angle, this is a trigonometric polynomial with frequencies $0$, $\pm2q$, $\pm4q$. For $\ell\in\{0,\pm2q,\pm4q\}$, the average of $e^{i(\ell+2q)\varphi_k}$ over the $4q$ grid points is zero unless $\ell=-2q$.
Thus the discrete average extracts only that coefficient, giving $m_qe^{2iq\theta}$.

\emph{Second-moment bound.}
(b) Write $x=\norm xu(\psi_x)$ and
$z=\tilde v\sqrt{q/2}\norm x$. The Jacobi--Anger expansion shows that the discrete average retains $J_{2qj}(z)$ with $j\equiv-1\pmod4$, each with a coefficient of modulus one.
The absolute values of these indices are $2q,6q,10q,\ldots$, each occurring once.
Since $|J_{-n}(z)|=|J_n(z)|$, we obtain $|T_{\rm cf}(x)|\le\sum_{j\ge1\ \mathrm{odd}}|J_{2qj}(z)|$. Let $I_0(r)=(2\pi)^{-1}\int_{-\pi}^{\pi}e^{r\cos t}\,dt$ denote the modified Bessel function of order zero.
The contour shift provides geometric decay in the Bessel index, so the retained terms can be summed.
Moving the contour in $J_n(z)=(2\pi)^{-1}\int_{-\pi}^{\pi}e^{i(z\sin t-nt)}dt$ to $t-i\vartheta$ gives $|J_n(z)|\le e^{-n\vartheta}I_0(|z|L_{\xi,\vartheta})$.
Summing over positive odd indices therefore yields
$|T_{\rm cf}(x)|\le I_0(c\norm x)e^{-2q\vartheta}/(1-e^{-4q\vartheta})$, where $c=v\sqrt{q/2}L_{\xi,\vartheta}$.

To bound the remaining factor, write $I_0(c\norm x)=(2\pi)^{-1}\int_0^{2\pi}e^{cu(t)^Tx}\,dt$.
Then $\E[I_0(c\norm X)^2]$ is an average of $\E e^{\lambda^TX}$ over vectors $\lambda=c[u(t)+u(s)]$.
All such vectors satisfy $\norm\lambda\le2c$. By symmetry, the moment generating function along
each ray is the expectation of a hyperbolic cosine and is nondecreasing with the radius. For real $\lambda$,
\eqref{s:PB} gives $\E e^{\lambda^T X}=e^{\norm\lambda^2}\Phi_x(\lambda)$ because the Gram--Charlier
quartic factor is unchanged under $t\mapsto it$. On the boundary $\norm\lambda=2c$, apply
Lemma~\ref{s:lem:circle} at the real radius parameter $2vL_{\xi,\vartheta}$. Since $|U|<1$,
$|1+U^2-2U\cos\Psi|^2\le16$, and the moment generating function there is bounded by
$16e^{q\tilde\Lambda(2vL_{\xi,\vartheta})}$.

\emph{Phase error and parameter bounds.}
(c) If $|\hat T-\E\hat T|\le|m_q|/\sqrt2$, the angle error satisfies
$d_q(\hat\theta,\theta)\le\pi/(8q)$. Chebyshev's inequality gives
$\Pr_\theta\{ |\hat T-\E_\theta\hat T|>|m_q|/\sqrt2\}\le2\bar V_q/(n|m_q|^2)\le1/4$.

(d) The required lower bound on the signal amplitude follows from the reverse triangle inequality:
\[
 |U_1(1+U_2^2)+U_2(1+U_1^2)|
 \ge |U_1|(1-|U_2|^2)-|U_2|(1+|U_1|^2)\ge |U_1|/2.
\]
Hence $|m_q|\ge e^{q\ell_\xi}/2$.
\end{proof}

For each fixed admissible triple, $\log n^*(q)\le\Gamma_a(v,\xi,\vartheta)q+O(1)$ as $q\to\infty$.
Define
\[
 \Gamma_a=\inf_{\substack{v,\vartheta>0,\ |\xi|<\pi/4\\ |w_1|>|w_2|}}\Gamma_a(v,\xi,\vartheta).
\]
Taking the limsup for each fixed triple and then the infimum gives $\limsup_{q\to\infty}q^{-1}\log n^*(q)\le\Gamma_a$. For each such triple the remaining conditions
in part (d) hold for all sufficiently large $q$.

For $a=4$, evaluate Proposition~\ref{s:prop:complex} at $(v,\xi,\vartheta)=(2.0899,0.1811,0.2641)$.
This gives $\Gamma_4(v,\xi,\vartheta)\simeq2.36884735$; numerical checks indicate that part (d)'s conditions hold for $q\ge9$.
Whenever those exact inequalities hold, the finite-$q$ guarantee is
\[
 n^*(q)\le512(1-e^{-4q\vartheta})^{-2}e^{q\Gamma_4(v,\xi,\vartheta)}+1.
\]
For any $0<a\le4$, multiplying this $v$ by $(4/a)^{1/4}$ leaves $a\tilde v^4$ and both roots $w_k$ unchanged, and provides a finite upper exponent. At $a=1$, the fixed triple $(1.64571694,0.32111649,0.47791005)$ gives $\Gamma_1(v,\xi,\vartheta)\simeq3.91127684$. These computations approximate bounds at fixed parameters;
they do not certify decimal endpoints or the parameter infimum. Combining the bounds gives
\[
 \varkappa_a\ \le\ \liminf_{q\to\infty}\frac{\log n^*(q)}q
 \le\ \limsup_{q\to\infty}\frac{\log n^*(q)}q
 \le\ \min\{2\varkappa_a,\Gamma_a\}.
\]

\paragraph{Conjecture.}
The existence of the limit of $q^{-1}\log n^*(q)$ is open. For each fixed $a\in(0,4]$, we conjecture convergence to $\Gamma_a$, equivalently the same limit for $q^{-1}\log(1/H_q^2)$. The radial and complex-frequency estimators extract the same angular harmonic. The unresolved step is to show that the optimized complex-frequency signal and second-moment
bounds lose no exponential factor relative to the radial estimator.
The following fits describe the behavior over a finite range of $q$.

\paragraph{Finite-range exponent fits.}
For each row of Table~\ref{tab:finite-fit}, we fit
$\log H_q^2=-\widehat\gamma q+c\log q+d$ by ordinary least squares on all integer $q$ in the indicated interval.
The fitted coefficients $\widehat\gamma$ are close to the fixed-parameter values of Proposition~\ref{s:prop:complex} reported above.
The table also gives the maximum absolute log residual. Comparing the overlapping windows checks
sensitivity to the fitting range. Because the windows overlap, they provide no independent asymptotic
validation; the limiting rate and subexponential correction remain open.
\begin{table}[!ht]
 \centering
 \caption{Finite-range fits to squared Hellinger distances; both windows end at $q=56$.
 Residuals are in natural-log units. The last column reports the numerical upper-bound evaluations
from Proposition~\ref{s:prop:complex} at the fixed parameters given above.}
 \label{tab:finite-fit}
 \begin{tabular}{rrrrrr}
 \toprule
 $a$ & $q_{\min}$ & $\widehat\gamma$ & $c$ & Max. log residual & \shortstack{Theoretical upper bound\\(numerically evaluated)} \\
 \midrule
 1 & 24 & 3.91120 & 0.4940 & $3.16\times10^{-5}$ & 3.91128 \\
 1 & 30 & 3.91121 & 0.4946 & $1.23\times10^{-5}$ & 3.91128 \\
 4 & 24 & 2.36864 & 0.4864 & $5.69\times10^{-4}$ & 2.3689 \\
 4 & 30 & 2.36873 & 0.4900 & $1.03\times10^{-4}$ & 2.3689 \\
 \bottomrule
 \end{tabular}
\end{table}

\paragraph{Computing the plotted distances.}\label{app:numerical-distances}
Equation~\eqref{eq:angular-density} reduces the integrals for the half-spacing pair to a radial
integral with an angular average. We construct the radial polynomials with $60+3q$ digits of working precision.
To avoid cancellation between nearly equal square roots, the angular Hellinger
integrand is evaluated in the equivalent form
$[2V/(\sqrt{U+V}+\sqrt{U-V})]^2$, with
$U=1+C(r)\cos(2t)$ and $V=B(r)\cos t$.
The radial integrands are multiplied by $e^{5q}$ for the squared Hellinger
distance and $M_q$, and by $e^{3q}$ for total variation during quadrature;
these factors are removed afterward.

We check convergence at $a\in\{0.5,1,2,3,4\}$ and $q\in\{3,10,20,56\}$.
The refined calculation uses $100+3q$ digits for the coefficients instead of $60+3q$, and $384$ angular nodes instead of $192$.
It also extends the radial cutoff in $u=r^2/2$ from $18q+40$ to $24q+60$. Table~\ref{tab:distance-check} shows two cases. Across all $20$ comparisons, the largest relative change in $H_q^2$ is $2.40\times10^{-15}$
(rounded up). The full distance data cover $q=3,\ldots,56$.

\begin{table}[!ht]
 \centering
 \caption{Radial refinement at $a=4$. Relative change is $|H^2_{\rm refined}-H^2_{\rm base}|/H^2_{\rm refined}$;
 the first row agrees at stored precision. These are consistency checks, not certified error bounds.}
 \label{tab:distance-check}
 \begin{tabular}{rrl}
 \toprule
 $q$ & Refined $H_q^2$ & Relative change \\
 \midrule
 10 & $4.145185221625136\times10^{-9}$ & No change at stored precision \\
 56 & $4.443402971679872\times10^{-56}$ & $2.04\times10^{-16}$ \\
 \bottomrule
 \end{tabular}
\end{table}

At $a=4$, independent Cartesian integration with product Hermite polynomials gives
$H_{10}^2\simeq4.145185221625239\times10^{-9}$.
The relative difference from the radial value is $2.48\times10^{-14}$ (rounded up).

\subsection{Normal sensitivity of a binary-source zero line}\label{s:binary-slope-rate}
In the binary-source example, the normal derivative measures how the CF changes across a zero line.
We now determine its exponential decay rate as the number of sources grows.
In the binary model of Section~\ref{sec:binary}, take $b>0$, $\tau\ge0$, and $b^2+\tau^2=1$.
Put $\omega_q(\nu)=\rho_0u(\theta)+\nu\up(\theta)$ and $S_q(\nu)=\|\nabla_\omega\Phi_x(\omega_q(\nu))\|$.
For any fixed $L\ge0$,
\begin{equation}
 \lim_{q\to\infty}\frac{1}{q}\log\sup_{|\nu|\le L} S_q(\nu)
 =-\gamma_{\rm bin}-\frac{\pi^2\tau^2}{16b^2},\qquad
 \gamma_{\rm bin}=-\frac{1}{\pi}\int_0^\pi
 \log\cos\!\left(\frac{\pi}{2}\cos\varphi\right)\,d\varphi .
 \label{s:eq:binary-slope-rate}
\end{equation}
The same limit holds at $\nu=0$, the point on the target line nearest the origin. More precisely,
$\log S_q(0)=-q(\gamma_{\rm bin}+\pi^2\tau^2/(16b^2))+\tfrac32\log q+O(1)$.

\begin{proof}
On the target zero line, differentiating any other factor leaves the target factor equal to zero.
Differentiating the target factor at its first zero therefore gives
\[
 S_q(\nu)=b\sqrt{2/q}\,e^{-\tau^2(\rho_0^2+\nu^2)/2}
 \prod_{j=1}^{q-1}
 \left|\cos\left(\frac{\pi}{2}\cos\varphi_j+
 b\sqrt{2/q}\,\nu\sin\varphi_j\right)\right|,
 \qquad \varphi_j=j\pi/q.
\]
At $\nu=0$, the cosine is positive between the endpoints and has a quadratic zero at each endpoint.
We separate these endpoint zeros from the smooth part before estimating the logarithm of the product.
The function $h(\varphi)=\log[\cos((\pi/2)\cos\varphi)/\sin^2\varphi]$ extends smoothly to both endpoints,
with value $\log(\pi/4)$. Its sum over the grid differs from $q/\pi$ times its integral by $O(1)$.
Using $\prod_{j=1}^{q-1}\sin\varphi_j=q/2^{q-1}$ and $\int_0^\pi\log\sin\varphi\,d\varphi=-\pi\log2$
gives $-q\gamma_{\rm bin}+2\log q+O(1)$ for the log product. Together with the Gaussian factor and
$b\sqrt{2/q}$ this proves the expansion at $\nu=0$.

For the uniform upper bound over $|\nu|\le L$, we must also allow perturbations that move cosine factors through zero.
We therefore truncate the logarithm before passing to a Riemann integral.
Fix $K>0$ and use $f_K(t)=\max\{\log|\cos t|,-K\}$, assigning the value $-K$ at zeros of $\cos t$.
The truncated function is continuous, even at the zeros of the cosine. The argument perturbation is uniformly at most $b\sqrt{2/q}L$.
For fixed $K$, uniform continuity of $f_K$ and convergence of the Riemann sums imply that the average of the truncated terms converges, uniformly over $|\nu|\le L$, to
$\pi^{-1}\int_0^\pi f_K((\pi/2)\cos\varphi)\,d\varphi$.
Let $K\to\infty$ to remove the truncation. Monotone convergence applied to the negatives of the
truncated terms bounds the limsup of the log product divided by $q$ by $-\gamma_{\rm bin}$.
The Gaussian factor contributes at most $-\pi^2\tau^2/(16b^2)$ to the limit, and $\nu=0$ gives the matching lower bound.
\end{proof}

Integrating the smooth remainder $h$ gives $\gamma_{\rm bin}\simeq1.50606$. At $b^2/\tau^2=10$
the Gaussian contribution is about $0.06169$. Equation~\eqref{s:eq:binary-slope-rate} gives a rate for
the largest normal slope on a bounded segment and allows individual offsets to have smaller slopes.

%% file: phase_details.tex
\subsection{Finite-sample illustration}\label{app:phase}

\paragraph{Simulation protocol.}
We use $a=4$ and $q\in\{3,4,5\}$. For each $q$ and each $x\in\{0.05,0.15,0.5,1.5,4\}$,
we round $x/H_q^2$ to the nearest positive integer to obtain $n$.
Figure~\ref{fig:phase} plots the resulting $nH_q^2$.
We draw $400$ independent samples of size $n$ at each pair $(q,n)$ and evaluate both estimators
on each sample.
At $\theta=0$, the unweighted statistic has mean $\E_0[e^{2iq\psi}]=\int_0^\infty A_1(\sqrt{2u})e^{-u}\,du$.
This mean is positive at all three values of $q$.

To sample $\phi_4$, we draw from the proposal with density
$\phi(s)(s^4+9)/12$ and accept with probability $(s^2-3)^2/(s^4+9)$.
The proposal is a mixture, with weights $3/4$ and $1/4$, of a standard normal variable and a chi variable with five degrees of freedom multiplied by an independent equiprobable sign.

We record the sample sizes, seed, success counts for each estimator, and counts of paired outcomes
on which the estimators disagree. Pointwise nominal $95\%$ Wilson score intervals
\citep{wilson1927,browncaidasgupta2001} quantify the Monte Carlo uncertainty of each estimator's success proportion.

%% file: ext_s2.tex
\section{Refining a labeled mixing column using characteristic-function zeros}\label{app:s2}
We refine a labeled mixing column using a known real zero of its source characteristic function and a
search window that isolates the target. We first identify the target line
(Section~\ref{sec:immunity}), then give conditions for a unique minimum using a finite frequency grid
(Section~\ref{sec:cfz-grid}), and finally derive the local covariance (Section~\ref{sec:cfz-precision}).
Table~\ref{tab:statistical-targets} compares the targets, supplied information, and limits used in the
paper. The source family here differs from the Gram--Charlier family in Theorem~\ref{thm:hellinger},
whose characteristic function is strictly positive on the real axis.

\begin{table}[!ht]
 \centering
 \caption{Overview of the statistical targets, supplied information, and limiting regimes.}
 \label{tab:statistical-targets}
 \begin{tabular}{@{}>{\raggedright\arraybackslash}p{0.14\linewidth}>{\raggedright\arraybackslash}p{0.33\linewidth}>{\raggedright\arraybackslash}p{0.20\linewidth}>{\raggedright\arraybackslash}p{0.25\linewidth}@{}}
 \toprule
 Analysis & Source family and supplied information & Target & Scope and limit \\
 \midrule
 Theorems~\ref{thm:hellinger}--\ref{thm:barrier} & Gram--Charlier; $q$, source law, spacing and column norm known & Common rotation modulo $\pi/q$ & Global, finite $n,q$; exponents at fixed $a$, $q\to\infty$ \\
 Appendix~\ref{app:s2} & Independent sources; known target real zero; fixed or independent isolating window; unique grid minimum & Labeled direction and zero-line offset & Local; fixed model and grid, $n\to\infty$ \\
 Appendix~\ref{app:s3} & Two sources under Assumption~A; target label known; source laws, scales and second direction unknown & Labeled direction & Local risk; fixed finite jump paths before $n\to\infty$, $t=n^{-1/2}$ \\
 Appendix~\ref{app:s4} & Known Gram--Charlier law and equal geometry; one complex zero and an isolating angle interval & Common rotation near the true angle & Local; first $n\to\infty$ at fixed $q$, then variance coefficient as $q\to\infty$ \\
 \bottomrule
 \end{tabular}
\end{table}

\subsection{The target line and the search region}\label{sec:immunity}
Write $X=a_1s_1+\eta$, where $a_1=r_1u(\theta_1)$ is the target column and $r_1>0$.
The aggregate interference $\eta$ is independent of the target symbol $s_1$.
For finitely many mutually independent sources with characteristic functions (CFs) $\chi_j(t)=\E e^{its_j}$,
\begin{equation}
 \Phi_x(\omega)=\prod_j\chi_j(a_j^T\omega).
 \label{eq:product}
\end{equation}

\paragraph{Moment invariance under independent interference.}
A source zero yields a moment that remains zero after independent interference is added.
Take $c^Ta_1=1$ and $d^Ta_1=0$. The first projection retains the target symbol with unit coefficient;
the second removes it. If $\E h(s_1+t)=0$ for every real $t$, then
\begin{equation}
 \E[h(c^TX)g(d^TX)]=0
 \label{eq:immune}
\end{equation}
whenever the product is absolutely integrable. Indeed, by independence and Fubini's theorem, integrating the product over $s_1$ first gives
$g(d^T\eta)\E_{s_1}[h(s_1+c^T\eta)]=0$ at every shift, for almost every $\eta$.
For a real zero $\omega_0$ of $\chi_1$, take $h(s)=e^{i\omega_0s}$, which has zero mean at every shift, and any bounded $g$.
Such a nonzero bounded $h$ obstructs bounded completeness of the location family \citep{mattner93,hushiu}; completeness in larger function classes requires separate conditions \citep{mattner92,andrews2017}. A related CF-zero cancellation removes input dependence in subtractively dithered quantization under nonoverload conditions \citep{schuchman,rapp}.

Parametrize a line by $\omega(\theta,\rho,\nu)=\rho u(\theta)+\nu\up(\theta)$, with $\rho>0$.
Its offset from the origin is $\rho$, and $\nu$ measures position along the line. The offset $\rho$ is
distinct from the half-spacing angle $\varrho$ in the rotation model.
A known positive real zero $\omega_0$ of $\chi_1$ fixes the relation $\rho_1=\omega_0/r_1$
between the target offset and the unknown amplitude $r_1$.

\begin{figure}[!tb]
 \centering
 \includegraphics[width=\linewidth]{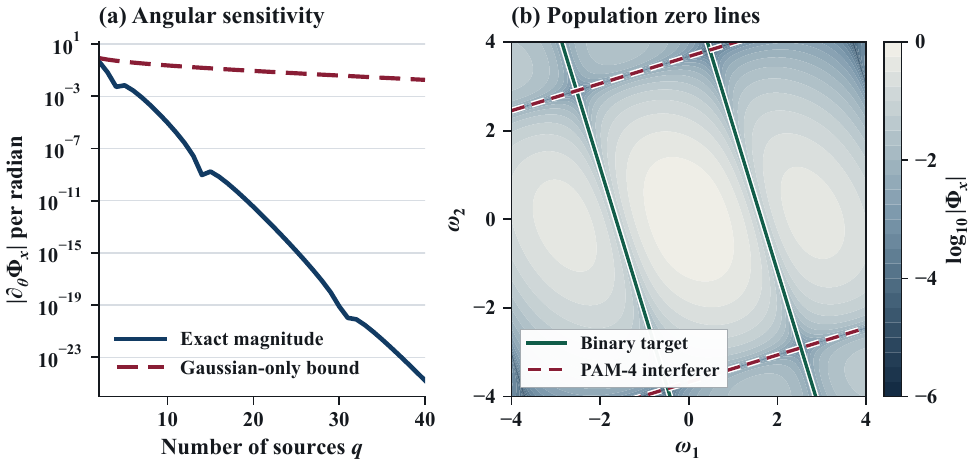}
 \caption{\textbf{(a)} Magnitude of the angular derivative at the first zero line of the equally spaced
 noisy binary model, at $b^2/\tau^2=10$ and a fixed tangential offset $\nu=1$. The dashed curve is the
 bound from \eqref{eq:binaryslope}, which retains only Gaussian damping. The exact rate in
 \eqref{s:eq:binary-slope-rate} concerns the maximum normal slope over a bounded segment centered at
 the point nearest the origin. \textbf{(b)} Population CF magnitude for the labeled-target
 example with Gaussian, four-level pulse-amplitude-modulated (PAM-4) and impulsive interference.
 The marked lines are the first positive and negative zeros from the binary target and the PAM-4
 interferer. The color scale is floored at $10^{-6}$ for display.}
 \label{fig:estimation}
\end{figure}

Using $c=u(\theta_1)/r_1$, $d=\up(\theta_1)$ and $g(z)=e^{i\nu z}$ in \eqref{eq:immune}
shows that the observed CF vanishes along $\omega(\theta_1,\rho_1,\nu)$.
Other sources can also contribute zero lines, as the binary target and discrete interferer do in
Figure~\ref{fig:estimation}(b). The following restrictions on the search region identify the line
associated with the labeled target.

\begin{corollary}[Identifying a target zero line]\label{cor:lines}
For mutually independent sources, the real zero set of $\Phi_x$ is the union of the hyperplanes
$a_j^T\omega=t$, where $t$ ranges over the real zeros of the corresponding source CF.
In two dimensions, assume that no source CF vanishes on an open interval. Work in a local chart for
directions modulo $\pi$, with $\rho>0$. Let the search region $\mathcal R$ contain $(\theta_1,\rho_1)$
and exclude the directions of other sources with real zeros. Assume that it also excludes every other
offset $\rho$ satisfying $\chi_1(\rho r_1)=0$. Then, for every $\varepsilon>0$, $(\theta_1,\rho_1)$ is the unique point in $\mathcal R$ whose candidate line satisfies $\Phi_x(\omega(\theta,\rho,\nu))=0$ for all $|\nu|<\varepsilon$.
\end{corollary}

\begin{samepage}
\begin{proof}
The union follows from \eqref{eq:product}. A finite union of closed sets with empty interior cannot cover an open interval. If the product vanishes along an interval, some factor must vanish on a subinterval. Its argument must be constant in $\nu$, so its column is parallel to the line normal. The angular and offset restrictions then select the target pair, using the fact that real CF zeros occur in sign pairs.
\end{proof}
\end{samepage}

In the equally spaced example, a connected search window containing the target and excluding both
neighboring directions places every candidate direction within one spacing of $\theta_1$. Within the
window we refine the direction and offset using characteristic-function moments evaluated on candidate
lines. The local precision depends on both the slopes of these moments and their covariance at a fixed
model size and a fixed number of frequency samples.

In Figure~\ref{fig:estimation}(a), the trial frequency is $\rho_0u(\theta')+\nu\up(\theta')$. At the true angle the angular derivative has magnitude $|\nu|$ times the normal slope because the characteristic function has zero tangential derivative on the line.

\subsection{Radial aliases for discrete sources}
Corollary~\ref{cor:lines} identifies candidate zero lines. Here we describe their offsets for the discrete source laws used in the zero-line example.

For an equiprobable binary target $s_1=\pm1$ with amplitude $r_1$, the first positive zero of
$\chi_1(\rho r_1)=\cos(\rho r_1)$ is $\rho_1=\pi/(2r_1)$. All positive zeros have offsets
$\rho=(2k+1)\rho_1$, with $k=0,1,\ldots$. Under the parametrization $\rho=\pi/(2r)$, these offsets
correspond to amplitudes $r_1/(2k+1)$. An amplitude range $[r_-,r_+]\ni r_1$ with $r_+<3r_-$
excludes all higher aliases ($k\ge1$).
A PAM-4 interferer with levels $\{\pm1,\pm3\}c$ has CF $\chi(t)=\cos(ct)\cos(2ct)$.
Its zero lines have normal $a_j$, so excluding $\theta_j$ from the search window excludes these lines.

\subsection{Identification and consistency with finitely many frequencies}\label{sec:cfz-grid}
The characteristic-function zero-line estimator (CFZ) uses the known $\omega_0$ and a compact
angle--offset region $\mathcal R$. It replaces the whole-line condition by sampled CF moments,
as in empirical characteristic-function inference \citep{feuervergermureika1977,feuervergermcdunnough1981}.
The frequency grid is finite; the search over $\mathcal R$ can be continuous. Write
$\hat\Phi_x(\omega)=n^{-1}\sum_i e^{i\omega^T X_i}$ and set
\[
 N(\theta,\rho)=\sum_{k=-L}^{L}|\hat\Phi_x(\omega(\theta,\rho,k\hat v/L))|^2 ,
\]
where the fixed integer $L\ge1$ determines the number of grid points and $\hat v>0$ estimates the
half-width $v$. In this appendix, $v$ measures width along the line; it denotes a frequency radius in
Appendix~\ref{app:s1}. CFZ minimizes
\begin{equation}
 R(\theta,\rho)=\frac{2N(\theta,\rho)}{D(\theta,\rho)},\qquad
 D(\theta,\rho)=N(\theta,(1-\delta)\rho)+N(\theta,(1+\delta)\rho)
 \label{eq:R}
\end{equation}
for a fixed relative offset $\delta\in(0,1)$. The denominator uses two lines flanking the candidate
line to account for CF decay away from the origin. Even when Corollary~\ref{cor:lines} identifies a
unique complete zero line, a different candidate can have every grid point at a crossing with other
zero lines. Finite-grid identification must exclude this possibility.

\emph{Identification.}
The required extra condition is that no nontarget candidate has all of its grid moments equal to zero.
Let $\xi=(\theta,\rho)$ and $\xi_0=(\theta_1,\rho_1)$. Write $N_0,D_0$ for the population versions of
$N,D$ in \eqref{eq:R} at the limiting width $v>0$. Suppose $D_0>0$ on the compact search region
$\mathcal R$. Under Corollary~\ref{cor:lines}, it suffices to show that every nontarget segment meets fewer than
$2L+1$ source zero lines. Each crossing covers at most one grid point, so at least one moment is
nonzero on every nontarget segment. All moments vanish at $\xi_0$; hence $N_0$ vanishes only there,
and $R_0=2N_0/D_0$ has a unique minimum at the target.

\emph{A grid alias.} Let two independent equiprobable binary sources have columns $a_1=(1,0)^T$ and
$a_2=(2,1)^T$. An independent Gaussian source with column $(1,-1)^T$ makes the mixture overcomplete
and adds no CF zeros. Take offsets $\rho\in[\pi/5,3\pi/5]$ and a sufficiently small angle window around
$\theta_1=0$ that excludes the binary interferer's direction. The only complete zero line in this
region is $(0,\pi/2)$. Choose the fixed width $v=L\pi$, so that the grid points are $\nu_k=k\pi$ for $k=-L,\ldots,L$. At the nontarget candidate $(0,\pi/4)$, every grid point satisfies
$a_2^T\omega(0,\pi/4,\nu_k)=\pi/2+k\pi$, so its grid energy is also zero.
Nevertheless, for $\delta=0.2$ the population denominator is positive throughout the offset interval
at $\theta=0$. To see this, take $\nu=0$. The smaller offset $0.8\rho$ has a nonzero target factor;
its interferer factor vanishes only at $\rho=5\pi/16$. At that offset, the larger frequency
$1.2\rho=3\pi/8$ gives two nonzero factors. By continuity and compactness, positivity
persists on a sufficiently small angle window. Thus a positive denominator and continuous-line
identification do not by themselves ensure a unique grid minimum.

\emph{Consistency.} Uniform convergence transfers the unique population minimum to consistent
sample minimizers. Treat the search region as fixed; when it is random, condition on the independent
side information that supplies it. For fixed $L$, the empirical CF
converges uniformly on compact frequency sets by the uniform law of large numbers for bounded continuous
parametric moments \citep[Lemma~2.4]{neweymcfadden}. If $\hat v\to_P v>0$, continuity of the population
CF then gives uniform convergence of the grid energies on $\mathcal R$. The positive minimum of $D_0$
ensures that the sample denominator is uniformly positive with probability tending to one. The sample
criterion converges uniformly to $R_0$, so a unique population minimum implies consistency of exact
global minimizers and of approximate global minimizers with objective gap $o_P(1)$.

\subsection{Moment sensitivity and asymptotic covariance}\label{sec:cfz-precision}
Local precision depends on the derivatives of the grid moments with respect to angle and offset,
and on the covariance of their sample averages. The next proposition gives the derivative at the
target and the asymptotic covariance for a consistent estimator. The latter result keeps the
observation law and grid fixed and specifies the required objective and stationarity rates.

\begin{proposition}[Moment sensitivity and asymptotic covariance]\label{prop:variance}
In two dimensions, suppose $s_1$ is independent of $\eta$. Assume that $\chi_1$ has a simple positive
real zero $\omega_0$: $\chi_1(\omega_0)=0$ and $\chi_1'(\omega_0)\ne0$.
Write $m(\theta,\rho;\nu)=\Phi_x(\omega(\theta,\rho,\nu))$ and $\xi_0=(\theta_1,\rho_1)$.
At $\xi_0$ the derivative row is
\[
 (\partial_\theta m,\partial_\rho m)=c(\nu)(-\nu,1),\qquad
 c(\nu)=r_1\chi_1'(\omega_0)\Phi_\eta(\omega(\theta_1,\rho_1,\nu)),
\]
where $\Phi_\eta$ is the CF of the aggregate interference.

For the limiting covariance of CFZ, let $n\to\infty$ with the observation law and $L$ fixed.
Assume $\xi_0\in\operatorname{int}\mathcal R$ and $D_0(\xi_0)>0$. Require $\E\|X\|<\infty$ and
$\hat v-v=O_P(n^{-1/2})$.
Stack the real and imaginary population grid moments at $\xi_0$ with width $v$.
Let their Jacobian $G$ have full column rank, and let $\Omega$ be the covariance of their
single-observation summands. For a consistent sequence $\hat\xi$
satisfying
\[
 R(\hat\xi)=O_P(n^{-1}),\qquad
 \|\nabla R(\hat\xi)\|=o_P(n^{-1/2}),
\]
with derivatives taken at fixed estimated width, $\sqrt n(\hat\xi-\xi_0)$ converges in distribution to
a centered normal with covariance
\[
 (G^TG)^{-1}G^T\Omega G(G^TG)^{-1}.
\]
These conditions on the estimator hold for an exact global minimizer if $D_0>0$ on $\mathcal R$ and
$N_0$ vanishes only at $\xi_0$. For a random search region the conclusion is conditional on the independent side information.
\end{proposition}

\begin{proof}
By independence, $m$ is the product of $\chi_1(a_1^T\omega)$ and $\Phi_\eta(\omega)$. At $\xi_0$ the first
factor vanishes. Its linear expansion and continuity of the second factor suffice to differentiate the
product there. Since $a_1^T\omega=r_1[\rho\cos(\theta-\theta_1)-\nu\sin(\theta-\theta_1)]$,
the row is $c(\nu)(-\nu,1)$. Two distinct $\nu_k$ with $c(\nu_k)\ne0$ give full column rank over
the reals.

We next account for estimating the grid width before linearizing the criterion.
Let $\bar m(\xi,v')$ stack the real and imaginary sample moments at width $v'$ and let $\bar M$ be its
derivative with respect to $\xi$. The derivative envelopes in $\xi$ and $v'$ are bounded by a constant
times $\|x\|$ on compact parameter and width sets. The first-moment assumption therefore gives local uniform
convergence of the derivatives. The population moments vanish at $\xi_0$ for every width, so their
population width derivative is zero there. Consequently
\[
 \sqrt n[\bar m(\xi_0,\hat v)-\bar m(\xi_0,v)]=o_P(1).
\]
The bounded fixed-grid moments satisfy the multivariate central limit theorem and $R(\xi_0)=O_P(n^{-1})$.
Under the global identification conditions an exact minimizer is consistent, lies in the interior with
probability tending to one, satisfies $R(\hat\xi)\le R(\xi_0)$, and has $\nabla R(\hat\xi)=0$.

At a consistent estimator satisfying the stated rates, the denominator derivative contributes only
a smaller-order term. Differentiating \eqref{eq:R} at fixed estimated width gives
\[
 \bar M(\hat\xi,\hat v)^T\bar m(\hat\xi,\hat v)
 =\tfrac14 D(\hat\xi)\nabla R(\hat\xi)+\tfrac14 R(\hat\xi)\nabla D(\hat\xi)
 =o_P(n^{-1/2}),
\]
since $D$ is bounded and $\nabla D=O_P(1)$ near the target. Expanding the moments along the segment
from $\xi_0$ to $\hat\xi$ and using uniform derivative convergence gives
\[
 (G^TG+o_P(1))(\hat\xi-\xi_0)
 =-G^T\bar m(\xi_0,v)+o_P(n^{-1/2}).
\]
Inversion and the central limit theorem give the stated covariance.
\end{proof}

\begin{figure}[!t]
 \centering
 \includegraphics[width=0.96\linewidth]{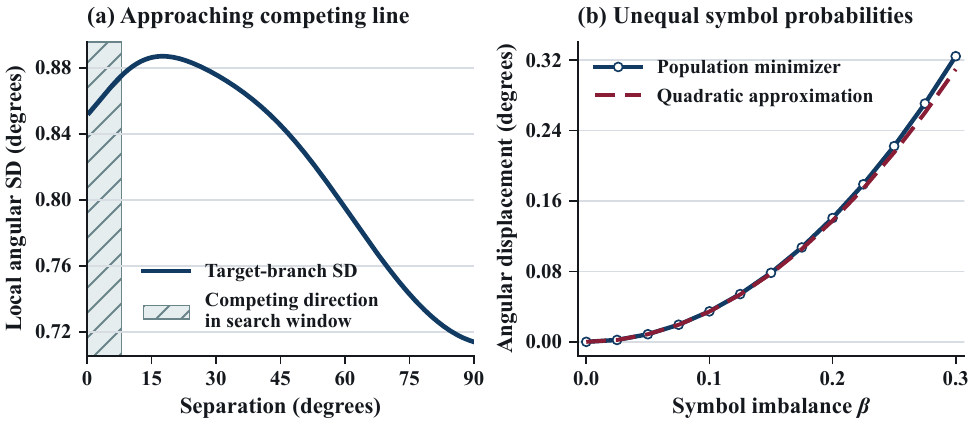}
 \caption{CFZ sensitivity calculated from the population distribution. \textbf{(a)} Predicted angular
 standard deviation as the PAM-4 direction approaches the target, with $n=10^4$ and per-interferer
 variance $P=0.5$. The label \emph{Target-branch SD} refers to the local minimizer near the target.
 Shading marks where the competing direction enters the search window. \textbf{(b)} Angular
 displacement of the population minimizer near the target as the binary symbols become imbalanced.
 The curve is compared with its quadratic approximation at $P=0.25$.}
 \label{fig:sensitivity}
\end{figure}

The local conclusion also applies to an exact minimizer on a smaller compact neighborhood of $\xi_0$
where the denominator is positive and the target is the unique population zero. Such a neighborhood
exists when $D_0(\xi_0)>0$, the moments are differentiable, and $G$ has full column rank. Figure~\ref{fig:sensitivity}(a) describes
the local solution near the target $\xi_0$. Applying the covariance formula to a numerical solution
requires the stated residual rate; a fixed optimization tolerance is insufficient.

The derivative and covariance calculations also apply to bounded functions of the projection
along the line. The following formulas cover these more general moments and weighted estimation.

\emph{Derivative for a general moment.} Assume $\E\|X\|<\infty$, and let $g$ and its derivative be
bounded. Define $m_g(\theta,\rho)=\E[e^{i\rho u(\theta)^TX}g(\up(\theta)^TX)]$.
Write $w=u(\theta_1)^T\eta$ and $z=\up(\theta_1)^T\eta$ for the two interference projections.
At the target pair,
\begin{equation}
 \begin{aligned}
 \partial_\theta m_g&=ir_1\chi_1'(\omega_0)\,\E[g'(z)e^{i\rho_1w}],\\
 \partial_\rho m_g&=r_1\chi_1'(\omega_0)\,\E[g(z)e^{i\rho_1w}].
 \end{aligned}
 \label{eq:jacobian}
\end{equation}

\emph{Covariance with a general weight.} Let $G$ be the Jacobian of the stacked real and imaginary population moments and $\Omega$ the covariance of their sample summands at the target. Under the usual local linearization for a consistent generalized method of moments estimator \citep{hansen1982,neweymcfadden}, the limiting covariance of the $\sqrt n$-scaled error is
\[
 K_W^{-1}G^TW\Omega WGK_W^{-1},\qquad K_W=G^TWG,
\]
where the limiting weight $W$ is symmetric positive definite and $G$ has full column rank.
For moments formed with $g_k$ and $g_l$, each covariance entry has absolute value at most
$\sup|g_k|\sup|g_l|$. These formulas use a fixed grid. Extending to continuum GMM requires
covariance-operator analysis \citep{carrascoflorens2000}.

\emph{Derivation.} Let $y_\theta=u(\theta)^TX$ and $z_\theta=\up(\theta)^TX$. Then
$\partial_\theta y_\theta=z_\theta$ and $\partial_\theta z_\theta=-y_\theta$, so for $g$ bounded with a bounded
derivative and $\E\norm{X}<\infty$, dominated convergence gives
\[
 \begin{aligned}
 \partial_\theta m_g&=\E\big[i\rho z_\theta e^{i\rho y_\theta}g(z_\theta)\big]-\E\big[y_\theta e^{i\rho y_\theta}g'(z_\theta)\big],\\
 \partial_\rho m_g&=\E\big[iy_\theta e^{i\rho y_\theta}g(z_\theta)\big].
 \end{aligned}
\]
At $(\theta_1,\rho_1)$, write $y=r_1s_1+w$. The target $s_1$ is independent of $(w,z)$. Finite first
moments of $s_1$ and $\eta$ follow from $\E\norm{X}<\infty$ by restricting the other independent term
to an event on which its norm is bounded. Conditioning on $(w,z)$ gives
\[
 \begin{aligned}
 &\E[e^{i\rho_1y}\tilde g(w,z)]\\
 &\quad=\chi_1(\omega_0)\E[e^{i\rho_1w}\tilde g(w,z)]=0,\\
 &\E[ye^{i\rho_1y}\tilde g(z)]\\
 &\quad=r_1\E[s_1e^{i\omega_0s_1}]\E[e^{i\rho_1w}\tilde g(z)]\\
 &\qquad+\chi_1(\omega_0)\E[we^{i\rho_1w}\tilde g(z)]\\
 &\quad=-ir_1\chi_1'(\omega_0)\E[e^{i\rho_1w}\tilde g(z)].
 \end{aligned}
\]
Here we used $\E[se^{i\omega s}]=-i\chi'(\omega)$. Substituting $\tilde g(w,z)=i\rho_1zg(z)$, $\tilde g=g'$ and $\tilde g=g$ gives \eqref{eq:jacobian}.

\emph{Moment covariance.} At the true parameter, let $g_k(X)=e^{i\rho_1y+i\nu_kz}$, where $y=u(\theta_1)^TX$ and $z=\up(\theta_1)^TX$. Then $\E g_k=0$, $|g_k|=1$, and
\[
 \E[g_k\overline{g_l}]=\E[e^{i(\nu_k-\nu_l)z}],\qquad
 \E[g_kg_l]=\chi_1(2\omega_0)\E[e^{2i\rho_1w+i(\nu_k+\nu_l)z}].
\]
When the source CFs are specified, both expectations can be evaluated using the product formula \eqref{eq:product}. To recover the covariance of the stacked real and imaginary parts, use
$\operatorname{Re}a\operatorname{Re}b=\frac12\operatorname{Re}(ab+a\bar b)$ and
$\operatorname{Im}a\operatorname{Im}b=\frac12\operatorname{Re}(a\bar b-ab)$.
The cross-covariance follows from $\operatorname{Re}a\operatorname{Im}b=\frac12\operatorname{Im}(ab-a\bar b)$.

\emph{Damping.} Each added independent interferer multiplies $c(\nu)$ by a factor of modulus at most one.
It can therefore reduce sensitivity while leaving the total variance of a single complex moment
unchanged. The real and imaginary parts of each $g_k$ have variances
$\frac12(1\pm\operatorname{Re}\E g_k^2)$, whose sum is $\E|g_k|^2=1$ for every interference law.
If $\Omega$ is positive definite and $W=\Omega^{-1}$, the first-order variance of $\hat\theta$ is at least
$\lambda_{\min}(\Omega)/(n\norm{G_\theta}^2)$.
Here $\norm{G_\theta}^2=\sum_k\nu_k^2|c(\nu_k)|^2$ measures sensitivity to the angle.

\subsection{A small imbalance in a binary target}\label{s:sub:imbalance}
The local covariance above assumes a target law with an exact real zero. We now examine how a small
binary imbalance changes the population criterion, keeping the grid, search region and interference
law fixed. Write $X=a_1s_1+\eta$, with $s_1$ independent of $\eta$.
Let $\Pr(s_1=\pm1)=(1\pm\beta)/2$, and denote the CF of $\eta$ by $\Psi$. Then
\[
 |\Phi_\beta(\omega)|^2=\big[\cos^2(a_1^T\omega)+\beta^2\sin^2(a_1^T\omega)\big]|\Psi(\omega)|^2.
\]
Thus the population numerator and denominator in \eqref{eq:R} are $N_\beta=N_0+\beta^2N_1$ and $D_\beta=D_0+\beta^2D_1$,
where $N_1,D_1$ use the sine-squared term on the corresponding lines.

\begin{lemma}[Local effect of binary imbalance]\label{s:lem:imbalance}
Suppose $D_0>0$ on a compact search region and $R_0=2N_0/D_0$ has a unique interior minimizer $\xi_0=(\theta_1,\rho_1)$
with positive definite Hessian $K_0=\nabla^2R_0(\xi_0)$. If $N_0,N_1,D_0,D_1$ are three times continuously differentiable near $\xi_0$,
then, for sufficiently small $|\beta|$, the population minimizer satisfies
\[
 \begin{aligned}
 \xi_\beta&=\xi_0-\beta^2K_0^{-1}\nabla Q(\xi_0)+O(\beta^4),\\
 Q&=\frac{2(N_1D_0-N_0D_1)}{D_0^2}.
 \end{aligned}
\]
\end{lemma}
\begin{proof}
Expanding the ratio gives $R_\beta=R_0+\beta^2Q+O(\beta^4)$. Apply the implicit function theorem to $\nabla R_\beta=0$ with parameter $\beta^2$.
Uniform convergence on the compact region keeps the minimizer near $\xi_0$.
\end{proof}
When the coefficient is nonzero, the population minimizer shifts by order $\beta^2$; the coefficient
may also vanish.

\subsection{Deterministic zero-line calculations}
Figure~\ref{fig:estimation}(b) uses a unit-amplitude binary target at $17^\circ$ and three independent
interferers. The Gaussian interferer at $62^\circ$ has variance $P$. The PAM-4 interferer at
$107^\circ$ has equiprobable levels $\{\pm1,\pm3\}\sqrt{P/5}$. The impulsive interferer at $152^\circ$
is zero with probability $0.95$ and otherwise Gaussian with variance $P/0.05$.
Independent Gaussian sensor noise has covariance $0.1I_2$. The figure uses $P=0.25$ and balanced
target symbols.

The sensitivity calculations use the angle window $[9^\circ,25^\circ]$ and amplitude interval $[0.5,2]$.
We set $L=6$ and $\delta=0.2$, and use width $v=1.349/\operatorname{IQR}(\up(17^\circ)^TX)$.
The projection uses the specified target direction and removes the target contribution. The
projected Gaussian mixture is symmetric and has a known CDF, so the population interquartile range
is twice the positive $0.75$ quantile. With this projection fixed, the density is positive and continuous at the
quartiles. The sample counterpart therefore has the width rate in Proposition~\ref{prop:variance}.

\emph{Grid identification at the original directions.}
For balanced symbols with the PAM-4 direction at $107^\circ$ and $P\in\{0.25,0.5\}$, the population
denominator is positive and the target is the unique grid minimum. The verification uses the intersection
count from Section~\ref{sec:cfz-grid}, followed by the denominator and rank checks.
First, the projected Gaussian sensor noise bounds the probability of any interval by the mass of a
centered noise interval of the same length. Writing $\Phi_{\rm N}$ for the standard normal CDF gives
$\operatorname{IQR}(\up(17^\circ)^TX)\ge2\sqrt{0.1}\,\Phi_{\rm N}^{-1}(0.75)$.
Consequently the grid segment has length $2v<6.34$. Transverse intersections with target zero lines are separated by at least
$\pi/\sin(8^\circ)>22$, and those with PAM-4 lines by at least
$\pi/(4\sqrt{P/5})>2.48$. A noncoincident segment can therefore meet at most one target line and
three PAM-4 lines. Gaussian and impulsive factors have no real zeros. A parallel candidate can
coincide only with the target's first zero line within the search window.

For the denominator, the two flanking offsets
have ratio $3/2$, so they cannot both coincide with target zeros, whose offsets are odd multiples
of $\rho_1$. At least one flank then has positive grid energy. For the rank condition at the target,
at most three grid points are PAM-4 zeros, leaving more than two nonzero rows at distinct $\nu_k$.

In Figure~\ref{fig:sensitivity}(a) we vary the PAM-4 direction and retain the balanced target.
The predicted local standard deviation rises from $0.714^\circ$ at $90^\circ$ separation to
$0.887^\circ$ at $18^\circ$ separation. The local calculation remains defined in the shaded range, where choosing the solution near the
target requires distinguishing it from competing solutions.

In Figure~\ref{fig:sensitivity}(b) we hold the original directions fixed and change the target's
symbol probabilities. Since the grid projection removes the target contribution, the population grid
stays fixed across these probabilities. The quadratic approximation in Lemma~\ref{s:lem:imbalance}
is $3.4414\beta^2$ degrees for this example. At $\beta=0.3$ the numerical minimizer near the target
is displaced by $0.3244^\circ$, illustrating a population shift that persists as $n$ grows.

\subsection{Annihilators from higher-order and complex zeros}
Real zeros supply the bounded exponential moments used above. Higher-order and complex zeros also
yield functions whose expectations vanish under every real shift, subject to the integrability
conditions below. These functions enter the information calculation in Appendix~\ref{app:s3}.
For a density $f$, let $\chi_f(\omega)=\int e^{i\omega s}f(s)\,ds$. Extend this definition to complex
$\omega$ wherever $\int e^{-\operatorname{Im}(\omega)s}f(s)\,ds<\infty$.
An admissible zero $\omega\in\C$ of multiplicity $m$ is defined by the following two conditions.
First, $s^je^{i\omega s}\in L^2(f)$ for $0\le j\le m-1$.
Second, $\chi_f^{(k)}(\omega)=0$ for $k<m$ and $\chi_f^{(m)}(\omega)\ne0$, with derivatives taken
along the real direction.

\begin{lemma}\label{s:lem:monomials}
If $\omega$ is an admissible zero of multiplicity $m$, then the functions $h_j(s)=s^je^{i\omega s}$
satisfy $\E h_j(s+d)=0$ for every $d\in\R$ and $0\le j<m$, where $s\sim f$. For the densities used in
Appendix~\ref{app:s3}, these functions belong to $\cE(f)$.
\end{lemma}

\begin{proof}
For each $d$ the polynomial expansion of $(s+d)^j$ and the assumed integrability give
\[
 \int h_j(s)f(s-d)\,ds
 =e^{i\omega d}\sum_{k\le j}\binom jk d^{j-k}(-i)^k\chi_f^{(k)}(\omega)=0.\qedhere
\]
\end{proof}

\emph{Integrability.} The moment identity in \eqref{eq:immune} requires
$\E|h(c^TX)g(d^TX)|<\infty$. For bounded $g$ and $h(s)=e^{i\omega_0s}$ at a real zero, this is automatic.
For $h(s)=s^je^{i\omega_0s}$, a sufficient condition at an admissible zero is
$\E[(1+|c^T\eta|^j)e^{-\operatorname{Im}(\omega_0)c^T\eta}]<\infty$, by independence and the
binomial formula. For a real zero this is a polynomial moment condition. For a complex zero the sign of
$\operatorname{Im}\omega_0$ determines which tail requires exponential control.

%% file: ext_s3.tex
\section{Local information in the rare-jump experiment}\label{app:s3}
This appendix studies local information about one labeled direction in a two-source model.
Consider the baseline $X=u(\theta_1)s_1+u(\theta_2)s_2$, with independent sources and distinct
directions modulo $\pi$. Both source laws and scales are unknown. The target is the labeled angle
$\theta_1$; the second angle $\theta_2$ is a nuisance parameter. We ask how much information remains
when rare independent shifts are also allowed, evaluating information at unit source variances.

A local path $\xi\in\mathcal T$ varies both angles and source laws and adds a fixed finite list of
independent jumps. At path coordinate $t\ge0$, jump $j$ has fixed displacement $d_j\in\R^2$ and
probability $t\lambda_j$, where $\lambda_j\ge0$ is fixed. Write $P_t^\xi$ for the one-observation law and
$\theta_{1,t}^\xi$ for its labeled first direction. The local scale in the risk bound is $t=n^{-1/2}$.
The proof first specifies the admissible changes in source laws and angles, then projects out the
nuisance scores \citep[Ch.~3]{bkrw1993}. Quadratic-mean expansions yield local Gaussian experiments
\citep[Ch.~7]{vdv1998}. We derive the risk bound from finite sets of feasible one-sided paths, so the
jump intensities remain nonnegative.

\subsection{Source assumptions and shift scores}
Let $f$ have mean $\bar s_f$ and variance $\sigma_f^2$. Its location score is
$\psi_f=-f'/f$, with information $J(f)=\int\psi_f^2f$.
We use the complex Hilbert space $L^2(f)$ with inner product $\langle u,v\rangle_f=\int u\bar vf$;
its mean-zero subspace is $L^2_0(f)=\{u\in L^2(f):\int uf=0\}$.
The scores are real, while the complex space also accommodates the Fourier functions used below.

\emph{Assumption A.} (i) $f>0$ on $\R$, $f$ is absolutely continuous, $\sigma_f^2<\infty$, $J(f)<\infty$ and
$\int s^2\psi_f^2f<\infty$.
\par(ii) $\int f(s-d)^2/f(s)\,ds<\infty$ for every $d\in\R$.
\par(iii) $\sup_{|t|\le\tau}\int f'(s-t)^2/f(s)\,ds<\infty$
for some $\tau>0$. Assumption A holds for the Gaussian law, $\phi_a$ with $0\le a<4$, finite Gaussian mixtures, the
logistic, Laplace and hyperbolic secant laws, Student $t$ laws with more than two degrees of freedom, and
constellations in Gaussian noise.

To identify the score components that survive the added shifts, we use functions whose expectation
vanishes under every translation of a source law. For a density $f$ satisfying Assumption~A(i)--(ii), define
\begin{equation}
 \cE(f)=\left\{h\in L^2(f):\int h(s)f(s-d)\,ds=0\ \text{for every }d\in\R\right\}.
 \label{eq:annihilator}
\end{equation}
By the shift-ratio condition, each translated expectation is a bounded linear functional on $L^2(f)$,
so $\cE(f)$ is closed.

To describe the score directions generated by shifts, define
$v_d(s)=f(s-d)/f(s)-1$. Under A(ii), these functions lie in $L^2_0(f)$. Let $\cU(f)$ be their closed span.
Since $\langle h,v_d\rangle_f=\int h(s)f(s-d)\,ds-\int hf$, a mean-zero function is orthogonal to
$\cU(f)$ exactly when its expectation vanishes under every translate of $f$. Therefore
$L^2_0(f)=\cU(f)\oplus\cE(f)$, with $\cE(f)$ defined in \eqref{eq:annihilator}.
The choice $d=0$ gives $\cE(f)\subset L^2_0(f)$. Because $f$ is real, $\cE(f)$ is closed under conjugation. The share
\[
 S(f)=\frac{\norm{\Pi_{\cE(f)}(s-\bar s_f)}_f^2}{\sigma_f^2}
\]
is the fraction of source variance in the component orthogonal to the shift-score space. It is invariant
under location and scale changes. Nuisance projection removes the component of the target source in
$\cU(f)$, leaving the variance fraction $S(f)$ in the information formula.

\begin{proposition}[Information retained under rare shifts]\label{prop:share}
Suppose both source laws satisfy Assumption~A. Let $\Delta=\theta_2-\theta_1$.
The efficient information for $\theta_1$ with and without the additional shifts is, respectively,
\[
 \tilde I=\frac{S(f_1)J(f_2)}{\sin^2\Delta},\qquad
 I=\frac{J(f_2)-1/J(f_1)}{\sin^2\Delta}.
\]
If $\tilde I>0$, every estimator sequence $\hat\theta_{1,n}$ satisfies
\[
 \sup_{\substack{F\subset\mathcal T\\0<|F|<\infty}}
 \liminf_{n\to\infty}\max_{\xi\in F}
 \E_{(P^\xi_{1/\sqrt n})^{\otimes n}}
 \!\left[n(\hat\theta_{1,n}-\theta^\xi_{1,1/\sqrt n})^2\right]
 \ge\frac{1}{\tilde I}.
\]
The bound also applies to any larger model containing these labeled local paths.
\begin{enumerate}
\item $S(f)\le1-1/(J(f)\operatorname{Var}(s))$.
\item $S(f)>0$ if $\chi_f$ has an admissible zero of finite multiplicity as defined in
Appendix~\ref{app:s2}.
\item $S(f)=0$ for Gaussian and Laplace laws.
\end{enumerate}
\end{proposition}

\subsection{Proof of the local information formula}
The proof proceeds in three steps. We establish a quadratic-mean expansion for simultaneous changes
in the angles, marginal laws, and jump probabilities. We then project out the nuisance scores and use
the remaining score to derive the risk bound for finite sets of feasible paths.

Write $\Delta=\theta_2-\theta_1$ and $A=[u(\theta_1),u(\theta_2)]$, so that $s=A^{-1}x$ and $\det A=\sin\Delta$.
\emph{Angle scores.} The log density is
$\log p(x)=-\log|\sin\Delta|+\log f_1(s_1)+\log f_2(s_2)$.
Using $\partial s/\partial\theta_1=-A^{-1}\up(\theta_1)s_1=(\cot\Delta\,s_1,-s_1/\sin\Delta)$ to differentiate it gives
\[
 \ell_{\theta_1}=\cot\Delta\,(1-s_1\psi_1(s_1))+s_1\psi_2(s_2)/\sin\Delta,
\]
and symmetrically $\ell_{\theta_2}=-\cot\Delta\,(1-s_2\psi_2(s_2))-s_2\psi_1(s_1)/\sin\Delta$, where
$\psi_i=\psi_{f_i}$. 

\emph{Local paths.} A path $\xi\in\mathcal T$ has
$\theta_{i,t}=\theta_i+t\gamma_i$ and $f_{i,t}=f_i(1+t b_i)$, where $\gamma_i\in\R$ and
$b_i$ is smooth, bounded, has bounded derivative, and satisfies $\int b_i f_i=0$. It adds
a finite list $(d_j,\lambda_j)_{j=1}^K$ of independent jumps with probabilities $t\lambda_j$
for $t\ge0$ sufficiently small. The angles stay in a common chart with distinct directions
modulo $\pi$; both source means and scales can vary. For small $t$, the tilted densities satisfy
Assumption~A. Bounded tilts preserve positivity and (ii). To verify (i) and (iii), differentiate
$f_i(1+t b_i)$ and use
\[
 \left\|\frac{f_i(\cdot-r)}{\sqrt{f_i}}\right\|_2
 \le 1+|r|\sup_{|z|\le\tau}
 \left\|\frac{f_i'(\cdot-z)}{\sqrt{f_i}}\right\|_2,\qquad |r|\le\tau,
\]
which follows from absolute continuity and Minkowski's inequality. Smooth bounded mean-zero
tilts are dense in the respective $L^2_0(f_i)$ spaces. Their score span closes to
$\mathcal M=(L^2_0(f_1)\otimes1)\oplus(1\otimes L^2_0(f_2))$, which includes the scale scores
$1-s_i\psi_i(s_i)$.

\emph{Joint angle and marginal perturbations.} We first verify quadratic-mean differentiability
without the jumps, allowing the mixing matrix and source densities to change together. Write
$g_t(s)=\sqrt{f_{1,t}(s_1)f_{2,t}(s_2)}$ and let $U_t$ be the pushforward isometry on
square roots,
\[
 (U_t g)(x)=|\det A_t|^{-1/2}g(A_t^{-1}x),\qquad
 A_t=[u(\theta_{1,t}),u(\theta_{2,t})].
\]
A change of variables shows that $U_t$ is strongly continuous on continuous compactly supported
functions. Isometry and density extend this property to $L^2$. The product square root
satisfies $g_t=g_0+\tfrac t2(b_1+b_2)g_0+o_{L^2}(t)$, where $b_i$ is evaluated at $s_i$.
For the angle derivative, A(i) gives
$\partial_i g_0$ and $s_j\partial_i g_0$ in $L^2$ for every $i,j$ by independence. On smooth
compactly supported $g$, differentiation gives
\[
 \frac{d}{dt}U_tg
 =U_t\!\left[-\tfrac12\operatorname{tr}(A_t^{-1}\dot A_t)g
            -\nabla g^{T}A_t^{-1}\dot A_t s\right].
\]
Its $L^2$ norm is bounded locally in $t$ by a constant times
$\|g\|_2+\sum_{i,j}\|s_j\partial_i g\|_2$.
Cutoff and then mollification approximate $g_0$ in this norm. The cutoff derivative
contributes only a bounded multiplier on the tails, and on each compact set the coordinate
weights are bounded. Integrating the uniform norm bound on difference quotients extends the derivative formula to $g_0$.
Combine this formula with the product expansion and strong continuity on $(b_1+b_2)g_0$.
The density before adding jumps, $p_t=(U_tg_t)^2$, then satisfies
\[
 \sqrt{p_t}=\sqrt p+\tfrac t2\ell_0\sqrt p+o_{L^2}(t),\qquad
 \ell_0(x)=\gamma_1\ell_{\theta_1}(x)+\gamma_2\ell_{\theta_2}(x)
                 +b_1(s_1)+b_2(s_2),
\]
where $p=p_0$ and the scores on the right use $s=A^{-1}x$.

\emph{Adding the jumps.} We now extend the expansion to the jump mixture as $t$ approaches zero
from above. Writing the square root of the mixture as a vector norm lets us control the shifted
components before differentiating.
Let $w_J(t)$ be the probability of the subset $J$ of jumps and
$d_J=\sum_{j\in J}d_j$. The full density is
$q_t=\sum_J w_J(t)p_t(\cdot-d_J)$. Write its square root as the pointwise Euclidean norm of
the vector with entries $\sqrt{w_J(t)}\sqrt{p_t(\cdot-d_J)}$. By the reverse triangle
inequality, replacing the shifted $\sqrt{p_t}$ by the correspondingly shifted $\sqrt p$ in all nonempty components
changes this norm in $L^2$ by at most
$(\sum_{J\ne\varnothing}w_J(t))^{1/2}\|\sqrt{p_t}-\sqrt p\|_2=O(t^{3/2})$.
The same inequality permits replacing the unshifted square root by its first-order
expansion with $o(t)$ error. The resulting norm, divided by $\sqrt p$, is
\[
 \sqrt{w_\varnothing(t)(1+t\ell_0/2)^2+\sum_{J\ne\varnothing} w_J(t)r_J},
 \qquad r_J(x)=\frac{p(x-d_J)}{p(x)} .
\]
Every $r_J$ belongs to $L^2(P_0)$ by A(ii) and independence in source coordinates.
The weights obey $w_\varnothing(t)=1-t\sum_j\lambda_j+O(t^2)$,
$w_{\{j\}}(t)=t\lambda_j+O(t^2)$, and $w_J(t)=O(t^2)$ for $|J|\ge2$. The displayed
norm has pointwise derivative $\tfrac12[\ell_0+\sum_j\lambda_j(r_{\{j\}}-1)]$ at zero.
To obtain the $L^2(P_0)$ derivative, first drop the term $t\ell_0/2$ inside the
first vector coordinate, which changes the norm by at most $t|\ell_0|/2$.
Then use $|\sqrt y-1|\le|y-1|$ for $y\ge0$. These bounds dominate the difference
quotient by $|\ell_0|/2+C(1+\sum_{J\ne\varnothing}r_J)$ for a finite path-dependent $C$.
Dominated convergence proves the joint one-sided quadratic-mean expansion.

\emph{Nuisance span.} We determine which parts of the angle score lie in the closed linear span of
the marginal and jump scores. The later risk argument enforces the nonnegative jump intensities.
For a single jump with $\delta=A^{-1}d$, the score per unit
intensity is
$v_{1,\delta_1}+v_{2,\delta_2}+v_{1,\delta_1}v_{2,\delta_2}$, where
$v_{i,a}=f_i(\cdot-a)/f_i-1$. The closed linear span of jump and marginal scores is
$\mathcal M\oplus(\cU(f_1)\hat\otimes\cU(f_2))$, where $\hat\otimes$ is the closed
Hilbert tensor product. Subtracting the additive marginal terms leaves every elementary
product $v_{1,a}v_{2,b}$ since the displacements are unrestricted.

\emph{Efficient score.} Project the target-angle score onto the orthogonal complement of this
nuisance space. Write $h_i=\Pi_{\cE(f_i)}(s_i-\bar s_{f_i})$.
The constant component of $s_1$, multiplied by $\psi_2$, is a marginal score.
The $\cU(f_1)$ component of $s_1$, multiplied by $\psi_2$, lies in the tensor nuisance space by Lemma~\ref{s:lem:psi}.
The residual is $\tilde\ell=h_1\psi_2/\sin\Delta$. Both factors have mean zero and $h_1\perp\cU(f_1)$, giving
orthogonality to the entire nuisance space above.

It remains to remove the second-angle nuisance direction. The projected second-angle score is $-h_2\psi_1/\sin\Delta$.
Its inner product with $\tilde\ell$ vanishes since $\langle h_1,\psi_1\rangle=0$. The remaining score therefore has information
$\tilde I=\|\tilde\ell\|^2=S(f_1)\sigma_{f_1}^2J(f_2)/\sin^2\Delta$ in this jump experiment.

Without jumps, removing the projection onto $\mathcal M$ leaves
$(s_1-\bar s_{f_1})\psi_2/\sin\Delta$ and $-(s_2-\bar s_{f_2})\psi_1/\sin\Delta$.
By Lemma~\ref{s:lem:psi}, their inner product is $-1/\sin^2\Delta$.
Projecting off the second residual therefore gives
$I=[J(f_2)-1/J(f_1)]/\sin^2\Delta$ at unit variances, in agreement with the classical semiparametric
ICA calculation \citep{amaricardoso,chenbickel}.

\emph{Local risk bound.} The scores of $\mathcal T$ form a convex cone. Angles
and marginal tilts can be added with signed coefficients, and finite lists of jumps
can be concatenated and their intensities multiplied by nonnegative constants.
For each $\xi\in\mathcal T$, the first-angle derivative is $\gamma_1$. When
$\tilde I>0$, it equals the inner product of the joint score of $\xi$ with
$\tilde\ell/\tilde I$. Indeed, this function has inner product one with
$\ell_{\theta_1}$ and zero with the second-angle and other nuisance scores.
It lies in the closed linear span of the tangent cone and has variance $1/\tilde I$.

We derive the risk bound on one-sided paths through Gaussian limit experiments;
see the finite-set formulation in \citet[Theorem~2]{vdv95}. Write $\mathcal C$ for the real feasible score cone
and $g=\tilde\ell/\tilde I$. We first work with the truncated loss $L_B(z)=\min\{z^2,B\}$,
then recover squared loss after passing to the Gaussian experiment.

First fix finitely many paths with scores $h_1,\ldots,h_m\in\mathcal C$ and first-angle
velocities $a_i=\langle g,h_i\rangle$. Let $\Lambda_{n,i}$ be their likelihood ratios relative to $P_0^{\otimes n}$.
The one-sided quadratic-mean expansions give joint convergence to
$\Lambda_i=\exp\{W(h_i)-\|h_i\|^2/2\}$. Here $W$ is a centered Gaussian process with
covariance $\E[W(h)W(k)]=\langle h,k\rangle$.
For any prior weights $\pi_i$, the maximum $L_B$-risk of any estimator is at least
\[
 \E_0\inf_{d\in\R}\sum_{i=1}^m\pi_i\Lambda_{n,i}L_B(d-a_i).
\]
Here $d=\sqrt n(\hat\theta_{1,n}-\theta_1)$, and the path centers are exactly
$\theta_1+a_i/\sqrt n$. The infimum can be restricted to the interval spanned
by the $a_i$. This infimum is a continuous function of the likelihood-ratio vector, bounded by
$B\sum_i\pi_i\Lambda_{n,i}$.
The likelihood ratios are uniformly integrable: they are nonnegative, have
expectation one, and their limiting lognormal variables also have expectation
one. Consequently the displayed expectation converges to the Bayes risk in
the corresponding finite Gaussian shift experiment.

Next we compare the feasible cone with a finite-dimensional Gaussian shift experiment.
Let $V$ be any finite-dimensional space spanned by feasible scores. Every
finite set $H\subset V$ can be translated into $\mathcal C$. Indeed, write its
elements as $h_i=\sum_j a_{ij}c_j$ in a basis of feasible scores $c_j$, and
choose $M_j\ge\max_i(-a_{ij},0)$. Then
$c=\sum_jM_jc_j\in\mathcal C$ and $h_i+c\in\mathcal C$ for every $i$.
Translating a Gaussian shift and its linear target by $c$ leaves its Bayes risk
unchanged. Thus finite subsets of $\mathcal C$ give at least the Bayes risks of
all finite priors in the Gaussian shift $Y=h+Z$ on $V$, with $Z\sim\mathcal N(0,I_V)$
and target $\langle g_V,h\rangle$, where $g_V=\Pi_Vg$.

To bound these risks from below, start with a Gaussian prior and approximate it by finite priors.
Under $h\sim\mathcal N(0,\tau^2I_V)$, the conditional distribution of the target is normal with
variance $\tau^2\|g_V\|^2/(1+\tau^2)$. Its conditional mean minimizes
$L_B$-risk, since each centered interval has maximal probability when centered
at the mean of a normal law. The Bayes risk is therefore
\[
 \E L_B\!\left(\frac{\tau\|g_V\|}{\sqrt{1+\tau^2}}Z_0\right),
 \qquad Z_0\sim\mathcal N(0,1).
\]
This risk is approximable by finite priors: truncate the prior to a compact
set, then quantize it. Truncation changes bounded risks by a quantity tending
to zero. Moving every support point by at most $\varepsilon$ changes risks
uniformly over decision rules by at most
$B\varepsilon/2+2\sqrt B\|g_V\|\varepsilon$, using the total-variation bound
between unit-covariance Gaussian shifts and the Lipschitz constant $2\sqrt B$
of $L_B$. Every such finite prior and its translated path set is fixed before
$n\to\infty$.

It remains to remove the prior and finite-dimensional approximations. First make the finite-prior
approximation arbitrarily accurate, then let $\tau\to\infty$, and finally take finite-dimensional
$V$ with $g_V\to g$. The supremum over finite path sets of the limiting lower $L_B$-risk is therefore
at least $\E L_B(\|g\|Z_0)$. The squared risk dominates this bound for every $B$.
Finally, monotone convergence as $B\to\infty$ gives
$\|g\|^2=1/\tilde I$, with exactly the quantifier order in
Proposition~\ref{prop:share}. The same restriction argument applies to any
larger model containing the same labeled local paths.

\subsection{Properties of the translate-annihilator space}
We now establish the projection properties used above. The location score belongs to $\cU(f)$,
which bounds the retained variance from above. An admissible characteristic-function zero gives a
function in $\cE(f)$ with nonzero covariance with the source, proving that the retained variance is positive.

\begin{lemma}\label{s:lem:psi}
Under Assumption A, $\psi_f\in\cU(f)$ and $\langle s-\bar s_f,\psi_f\rangle_f=1$. Hence $S(f)\le1-1/(\sigma_f^2J(f))$.
\end{lemma}

\begin{proof}
Let $h\in\cE(f)$ and set $q_d=[f(\cdot-d)-f]/(df)=-d^{-1}\int_0^df'(\cdot-t)/f\,dt$.
Assumption A(iii) and Minkowski's inequality bound these quotients in $L^2(f)$ for $|d|\le\tau$.
Absolute continuity and Lebesgue differentiation give $q_d\to\psi_f$ almost everywhere as $d\to0$. A bounded sequence that converges almost everywhere converges
weakly in $L^2(f)$ to the same limit, so $\langle h,\psi_f\rangle=\lim\langle h,q_d\rangle=\lim
d^{-1}\int h(s)[f(s-d)-f(s)]\,ds=0$. Hence $\psi_f\perp\cE(f)$, and $\psi_f\in L^2_0(f)$ gives $\psi_f\in\cU(f)$.
Integration by parts gives $\langle s-\bar s_f,\psi_f\rangle=1$, so by the Cauchy--Schwarz inequality
$\norm{\Pi_{\cU}(s-\bar s_f)}^2\ge1/J(f)$ and $S(f)\le1-1/(\sigma_f^2J(f))$.
\end{proof}

By the Cram\'er--Rao inequality $\sigma_f^2J(f)\ge1$, with equality only for the Gaussian law.

\begin{proposition}[Positivity]\label{s:prop:positive}
If $\chi_f$ has an admissible zero $\omega$ of finite multiplicity $m$, then $S(f)>0$.
\end{proposition}

\begin{proof}
By Lemma~\ref{s:lem:monomials}, $h=s^{m-1}e^{i\omega s}\in\cE(f)$, and
\[
 \begin{aligned}
 \langle s-\bar s_f,h\rangle_f&=\E[s^me^{-i\bar\omega s}]-\bar s_f\E[s^{m-1}e^{-i\bar\omega s}]\\
 &=(-i)^m\chi_f^{(m)}(-\bar\omega)=i^m\overline{\chi_f^{(m)}(\omega)}\ne0;
 \end{aligned}
\]
the second term vanishes because
$\chi_f^{(m-1)}(-\bar\omega)=(-1)^{m-1}\overline{\chi_f^{(m-1)}(\omega)}=0$. Hence $\Pi_{\cE(f)}(s-\bar s_f)\ne0$.
The $m$th derivative exists: the Cauchy--Schwarz inequality makes
$|s|^m|e^{i\omega s}|=|s|\,|s^{m-1}e^{i\omega s}|$ integrable.
\end{proof}

\begin{proposition}[Gaussian and Laplace laws]\label{s:prop:complete}
$\cE(f)=\{0\}$, hence $S(f)=0$, for the Gaussian and the Laplace law.
\end{proposition}

\begin{proof}
For a Gaussian density and $h\in\cE(\phi)$, the Weierstrass transform
$K(d)=\int h(s)\phi(s-d)\,ds$ vanishes identically. This transform is injective on $L^2(\phi)$ by
uniqueness of the two-sided Laplace transform of $h\phi$. Hence $h=0$. For a Laplace density $f=(2b)^{-1}e^{-|s|/b}$, $(1-b^2\partial^2)f=\delta$ in the sense of
distributions, so $h\in\cE(f)$ gives $h=(1-b^2\partial_d^2)(h\star f)=0$. Here the distributional convolution
identity follows by testing against compactly supported smooth functions and using
$f(s-d)/f(s)\le e^{|d|/b}$ to justify the exchange of integrals.
\end{proof}

More generally, $S(f)=0$ if and only if $s-\bar s_f$ is orthogonal to the full annihilator space.

\subsection{Closed form for Gram--Charlier laws}
For the strictly positive polynomial perturbations below, the roots of a polynomial determine the
finite-dimensional space $\cE(f)$. A Gram matrix then computes the projection defining $S(f)$.
\begin{theorem}\label{s:thm:gc}
Let $f=\phi\,g$, where $g=\sum_{k=0}^Mg_k\He_k$, $g_0=1$, $g_M\ne0$, and $g\ge c>0$.
Define $G(z)=\sum_kg_kz^k$. Then $\chi_f(\omega)=e^{-\omega^2/2}G(i\omega)$ and
$\E_fe^{zs}=M_f(z):=e^{z^2/2}G(z)$. If $G$ has the roots
$\lambda_\varrho$ with multiplicities $m_\varrho$, then $\cE(f)=\operatorname{span}\{s^je^{\lambda_\varrho s}:0\le
j<m_\varrho\}$, of dimension $M$. For simple roots, $S(f)=b^H\Gamma^{-1}b/\sigma_f^2$ with
$\Gamma_{ij}=M_f(\lambda_j+\bar\lambda_i)$ and $b_i=M_f'(\bar\lambda_i)$.
\end{theorem}

\begin{proof}
Since $g\ge c$, $L^2(f)\subset L^2(\phi)$. For $h\in L^2(f)$, $K(d)=\int h(s)\phi(s-d)\,ds$ is entire, because
$\int|h||\phi(\cdot-d)|\le\norm h_{L^2(\phi)}(\int|\phi(s-d)|^2/\phi(s)\,ds)^{1/2}<\infty$ for every $d\in\C$. Rodrigues' formula gives
$\He_k(s-d)\phi(s-d)=\partial_d^k\phi(s-d)$, hence $\int h(s)f(s-d)\,ds=G(D)K(d)$.
Therefore $h\in\cE(f)$ if and only if $G(D)K\equiv0$. The kernel of $G(D)$ consists of the exponential polynomials
$\sum_\varrho P_\varrho(d)e^{\lambda_\varrho d}$ with $\deg P_\varrho<m_\varrho$. The Weierstrass transform is injective on $L^2(\phi)$.
It maps $s^je^{\lambda s}$ to $e^{\lambda d+\lambda^2/2}$ times a monic polynomial of degree $j$ in $d$. Thus the preimage of the kernel is the stated span, which lies in $L^2(f)$. Finally
$\langle e^{\lambda_js},e^{\lambda_is}\rangle_f=\E e^{(\lambda_j+\bar\lambda_i)s}$ and
$\langle s,e^{\lambda_is}\rangle_f=\E[se^{\bar\lambda_is}]$, and the projection of $s-\bar s_f$ equals that of $s$
because constants are orthogonal to $\cE(f)$.
\end{proof}

\subsection{A 10 dB example}
For identical unit-variance source laws with $J=J(f)>1$, the fraction of information retained is
$\tilde I/I=S(f)/(1-1/J^2)$. Take $Y=mB+Z$, where $B$ is an equiprobable sign and $Z\sim\mathcal N(0,1)$ is independent.
Set $m^2=10$. To evaluate $J$, standardize $Y$ to unit variance; scale invariance allows us to compute
$S(f)$ in the original units. The source CF is
$e^{-\omega^2/2}\cos(m\omega)$, with zeros $\omega_k=(\pi/2+k\pi)/m$. Projection onto
$\{\sin(\omega_kY):0\le k<64\}$ gives a lower bound on $S(f)$, since these functions belong to
the full translate-annihilator space. Evaluating the sine Gram matrix from the CF and the location
information by quadrature gives $S(f)>0.9047$ and $\tilde I/I>0.9126$. The resulting information bound concerns the rare-jump experiment; attaining it with unknown source
laws and an unknown second direction remains open.

%% file: ext_s4.tex
\section{Local variance from one complex-frequency moment}\label{app:s4}
Empirical Fourier and moment-generating transforms provide a common estimation framework
\citep{feuervergermcdunnough1984}. We use one complex-frequency moment to relate sensitivity to local asymptotic variance.
First take $n\to\infty$ at fixed $q$; then study how the resulting variance coefficient grows with $q$.

Fix $0<a\le4$ in the Gram--Charlier rotation model of Theorem~\ref{thm:hellinger}. The target column is
$a_0=\sqrt{2/q}\,u(\theta)$. The source CF $\chi(t)=e^{-t^2/2}(1+at^4/24)$ has the simple zero
$t_0=(24/a)^{1/4}e^{i\pi/4}$. For $z\in\C$, consider the sample mean of
\[
 \begin{aligned}
 Z(\theta')&=e^{i\omega(\theta')^TX},\\
 \omega(\theta')&=\sqrt{q/2}\,[t_0u(\theta')+z\,\up(\theta')].
 \end{aligned}
\]
Expectations are taken under the fixed true law $P_\theta$, while the candidate frequency varies with $\theta'$.
At the true angle, the target source factor gives $\E Z(\theta)=0$.
The source exponential moments ensure integrability at these complex frequencies.

To apply the product formula, write the frequency seen by each source as
$t(\varphi)=t_0\cos\varphi+z\sin\varphi$ and $t_j=t(j\pi/q)$.
The index $j=0$ denotes the target; the other indices denote the remaining equally spaced columns. For fixed $z\ne0$, assume $\chi(t(\varphi))\ne0$ for $0<\varphi<\pi$. Then
\[
 D_q=\left.\partial_{\theta'}\E Z(\theta')\right|_{\theta'=\theta}
     =-z\chi'(t_0)\prod_{j=1}^{q-1}\chi(t_j)\ne0.
\]
For example, $z=\pm it_0$ is excluded because $t(\pi/2)$ is another zero.

For fixed $q$, choose a small closed interval around $\theta$ on which it is the unique population root;
$D_q\ne0$ guarantees that such an interval exists. Minimize the squared modulus of the sample mean there.
The exponential tails give uniform laws of large numbers for the moment and its first two derivatives on this interval. Because the angle is real, the least-squares condition projects the complex sample fluctuation onto
the derivative direction. Expanding this condition to first order yields
$\sqrt n(\hat\theta-\theta)\Rightarrow\mathcal N(0,V_q)$ with
\[
 V_q=\frac{\E|Z|^2+\operatorname{Re}(e^{-2i\arg D_q}\E Z^2)}
               {2|D_q|^2},\qquad Z=Z(\theta).
\]
This expression uses both real and imaginary parts with equal weight. Write $M(u)=e^{u^2/2}(1+au^4/24)$.
Independence and symmetry give $\E|Z|^2=\prod_j M(2\operatorname{Im}t_j)$ and
$\E Z^2=\prod_j\chi(2t_j)$. Each factor satisfies $|\chi(2t_j)|\le M(2\operatorname{Im}t_j)$.
The target factor gives the uniform estimate
\[
 \frac{|\E Z^2|}{\E|Z|^2}\le
 \frac{|\chi(2t_0)|}{M(2\operatorname{Im}t_0)}
 =3e^{-\sqrt{24/a}}\le 3e^{-\sqrt 6}<1 .
\]
Thus the second complex moment cannot cancel the positive term $\E|Z|^2$ in the numerator:
$V_q$ is bounded above and below by fixed positive multiples of $\E|Z|^2/|D_q|^2$.

To determine the growth of the variance coefficient, we collect each nontarget source's contribution
to the logarithm of this ratio. Let
\[
 \begin{aligned}
 F_a(t)&=|t|^2+\log\big(1+\tfrac{2a}3(\operatorname{Im}t)^4\big)
                 -2\log\big|1+\tfrac a{24}t^4\big|,\\
 \gamma(z)&=\pi^{-1}\int_0^\pi F_a(t(\varphi))\,d\varphi .
 \end{aligned}
\]
Since $F_a(t)=\log M(2\operatorname{Im}t)-2\log|\chi(t)|$,
$\log V_q=\sum_{j=1}^{q-1}F_a(t_j)+O_{a,z}(1)$.
To approximate this sum by an integral, first remove the singularities at the endpoints.
The zeros at $\pm t_0$ are simple, with $t'(0)=z$ and $t'(\pi)=-z$.
Hence $F_a(t(\varphi))+2\log\sin\varphi$ extends smoothly to $[0,\pi]$.
For this smooth function, the uniform-grid sum differs from $q$ times the integral average by $O(1)$.
We account for the subtracted term exactly: $\prod_{j=1}^{q-1}\sin(j\pi/q)=q/2^{q-1}$ and
$\int_0^\pi\log\sin\varphi\,d\varphi=-\pi\log2$. It follows that
\[
 \log V_q=q\gamma(z)-2\log q+O_{a,z}(1).
\]
In the fixed-$q$ normal approximation, a standard error of $\pi/(4q)$ corresponds to
$n=V_q(4q/\pi)^2$. A growing-$q$ sample-size guarantee would require bounds uniform in both $n$ and $q$.

Choosing the frequency and combining several moments remain open problems beyond this calculation
on a fixed local interval.